\documentclass[10pt,twocolumn,letterpaper]{article}

\usepackage{myiccv}
\usepackage{my}

\usepackage{pdfpages}

\iccvfinalcopy 

\ificcvfinal\fi

\begin{document}

\title{Semantic Segmentation-Assisted Instance Feature Fusion for Multi-Level\\ 3D Part Instance Segmentation}

\author{
  {Chun-Yu Sun}$^{1,2}$\qquad
  {Xin Tong}$^{2}$\qquad
  {Yang Liu}$^{2}$  \\
  $^1${Tsinghua University}\qquad
  $^2${Microsoft Research Asia}  \\
  {\tt\small sunchyqd@gmail.com\qquad
            \{xtong,yangliu\}@microsoft.com}
}

\maketitle
\ificcvfinal\thispagestyle{empty}\fi

\begin{abstract}
Recognizing 3D part instances from a 3D point cloud is crucial for 3D structure and scene understanding. Several learning-based approaches use semantic segmentation and instance center prediction as training tasks and fail to further exploit the inherent relationship between shape semantics and part instances. In this paper, we present a new method for 3D part  instance segmentation. Our method exploits semantic segmentation to fuse nonlocal instance features, such as center prediction, and further enhances the fusion scheme in a multi- and cross-level way. We also propose a semantic region center prediction task to train and leverage the prediction results to improve the clustering of instance points. Our method outperforms existing methods with a large-margin improvement in the PartNet benchmark. We also demonstrate that our feature fusion scheme can be applied to other existing methods to improve their performance in indoor scene instance segmentation tasks. 
\end{abstract}

\section{Introduction} \label{sec:intro}

3D instance segmentation is the task of distinguishing 3D instances from 3D data at the object or part level and extracting the instance semantics simultaneously~\cite{tchapmi2017segcloud,yang2019learning,lahoud20193d,zhang2020instance}. It is essential for various applications, such as remote sensing, autonomous driving, mixed reality, 3D reverse engineering, and robotics. However, it is also a challenging task due to the diverse geometry and irregular distribution of 3D instances. Extracting part-level instances like chair wheels and desk legs becomes more difficult than segmenting object-level instances like beds and bookshelves, as the shape of the parts have large variations in structure and geometry, while part-annotated data are scarce.

A popular learning-based approach to 3D instance segmentation follows the encoder-decoder paradigm, which   predicts pointwise semantic labels and pointwise instance-aware features intercurrently~\cite{lahoud20193d,SASO,engelmann20203d,liu2020learning,jiang2020pointgroup,zhang2021,DyCo3D}. Instance-sensitive features can be either 3D instance centers, which have a clear geometric and semantic meaning, or feature vectors embedded in a high-dimensional space, where the feature vectors of the points within the same instance should be similar. The feature vectors of the points belonging to different instances are far apart from each other. Instance-aware features are used to group points into 3D instances via suitable clustering algorithms.   
Point semantics is usually used only in the clustering step. As the point set with the same semantics in a scene is composed of one or multiple 3D instances, it is natural to think about how to utilize this relation maximally. The work of \cite{ASIS} and \cite{zhao2020jsnet} associates semantic features with instance-aware features to improve the learning of semantic  features and instance features. However, they only fuse instance features with semantic features in a pointwise manner, without using semantics-similar points to provide nonlocal and robust guidance to instance features. 

In this study, we leverage the probability vectors of semantic segmentation to help aggregate the instance features of points in an explicit and nonlocal way. We call our approach \emph{semantic segmentation-assisted instance feature fusion}. The aggregated instance feature combined with the pointwise instance feature provides both global and local guidance to improve instance center prediction robustly, whose accuracy is critical to the final quality of instance clustering. Compared  to existing feature fusion schemes~\cite{ASIS,zhao2020jsnet}, our feature fusion strategy is more effective and simpler, as verified by our experiments. 

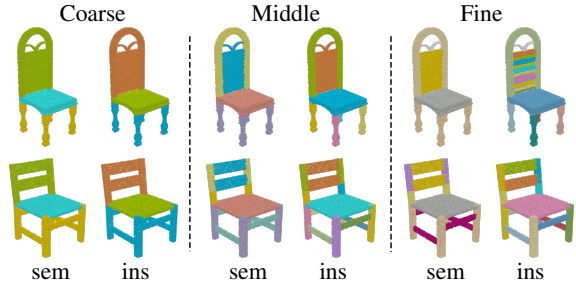
\begin{figure}[t]
\centering
  \begin{overpic}[width=0.9\linewidth]{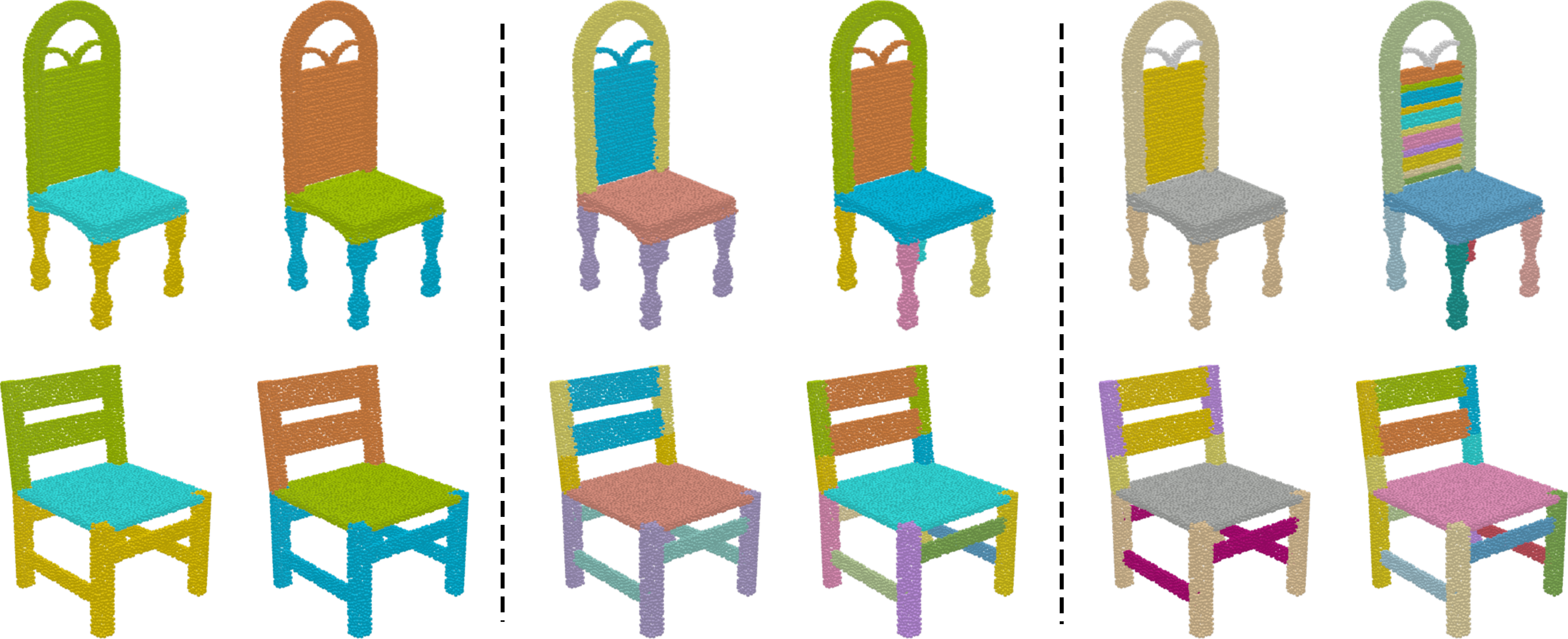}
    \put(4,-3.8){\small sem}
    \put(20,-3.8){\small ins}
    \put(39,-3.8){\small sem}
    \put(55,-3.8){\small ins}
    \put(74,-3.8){\small sem}
    \put(90,-3.8){\small ins}
    \put(9,42){\small Coarse}
    \put(43,42){\small Middle}
    \put(80,42){\small Fine}
   \end{overpic}
   \vspace{3mm}
  \caption{Illustration of 3D models with fine-grained and hierarchical part structures. Models are selected from PartNet~\cite{Mo2019}. From left to right: part semantics and part instances at the coarse, middle and fine level. Point colors are assigned to distinguish different part semantics and part instances. }
  \label{fig:partstructure} \vspace{-6mm}
\end{figure}

Human-made 3D shapes, such as chairs, are composed of a set of meaningful parts and exhibit hierarchical 3D structures (see \cref{fig:partstructure}). Extracting multi-level part instances from the point cloud is challenging, especially for fine-level 3D instances, such as chair wheels. Existing studies independently performed 3D part instance segmentation on each structural level and also suffered from the insufficient labeled-data issue on some shape categories. By utilizing the hierarchy of shape semantics and part instances, we extend our feature fusion scheme in a multi- and cross-level manner, where the probability feature vectors at all levels are used to aggregate instance features.  
Furthermore, to better distinguish part instances that are very close to each other, we propose to predict the centers of grouped instances, called \emph{semantic region centers}, and use them to push the predicted instance centers away from them, as the semantic region centers play the role of the centers of a group of semantics-same part instances. On the PartNet dataset~\cite{Mo2019} in which 3D shapes have 3-level semantic part instances, our approach exceeds all existing approaches on the mean average precision (mAP) part category (IoU$>$\num{0.5}) by an average margin of +\SI{6.6}{\percent} on 24 shape categories. 

Our semantic segmentation-assisted instance feature fusion scheme is simple and lightweight; it is not limited to 3D part instance segmentation and can be extended to 3D instance segmentation for indoor scenes. We integrated several state-of-the-art 3D instance segmentation frameworks with our feature fusion scheme and observed consistent improvements on the benchmark of ScanNet~\cite{dai2017scannet} and S3DIS~\cite{S3DIS}, which demonstrate the efficacy and generality of our approach.

\myparagraph{Contributions} We make two contributions to tackle 3D instance segmentation: (1) We propose an instance feature fusion strategy that directly fuses instance features in a nonlocal way according to the guidance of semantic segmentation to improve instance center prediction. This strategy is lightweight and easily incorporated into many 3D instance segmentation frameworks for both 3D object and part instance segmentation. (2) Our multi- and cross-level instance feature fusion and the use of the semantic region center are effective for multi-level part instance segmentation and achieve the best performance on the PartNet benchmark. Our code and trained models are publicly available at \url{https://isunchy.github.io/projects/3d_instance_segmentation.html}.

\section{Related Work} \label{sec:related}

\myparagraph{2D instance segmentation} As surveyed by \cite{Hafiz2020survey},  four typical paradigms exist in the literature. The methods in the first paradigm generate mask proposals and then assign suitable shape semantics to the proposals~\cite{fastrcnn,ren2015faster,wang2020solo}. The second one detects multiple objects using boxes and then extracts object masks within the boxes. Mask R-CNN~\cite{he2017mask} is one of the representative methods. The third is a bottom-up approach that predicts the semantic labels of each pixel and then groups pixels into 2D instances~\cite{bai2017deep}. Its computation is relatively heavy due to per-pixel prediction. The fourth paradigm suggests using \textit{dense sliding windows} techniques to generate mask proposals and mask scores for better instance segmentation~\cite{dai2016instance,chen2019tensormask}. For detailed surveys, see articles~\cite{Hafiz2020survey,Zhang2021survey,Minaee2021survey}.  

\myparagraph{3D Instance segmentation} The existing 3D approaches follow the paradigms of 2D instance segmentation (\cf surveys~\cite{Guo2019Survey,he2021deep}). \textit{Proposal-based methods} ~\cite{Mo2019,Jiang2020withoutdetection} predict a fixed number of instance segmentation masks and match them with the ground truth using the Hungarian algorithm or a trainable assignment module. The learned matching scores are used to group 3D points into instances.  \textit{Detection-based methods}~\cite{hou20193d,yi2019gspn,yang2019learning,engelmann20203d} generate high-objectness 3D proposals like boxes and then refine them to obtain instance masks. 

\textit{Clustering-based methods} first produce per-point predictions and then use clustering methods to group points into instances. SGPN~\cite{SGPN} predicts the similarity score of any two points and merges points into instance groups according to the scores. MASC~\cite{liu2019masc} predicts the multiscale affinity between neighboring voxels, for instance, clustering. Hao \etal~\cite{han2020occuseg} regress the instance voxel occupancy for more accurate segmentation outputs. PointGroup~\cite{jiang2020pointgroup} uses both the original and offset-shifted point sets to group points into candidate instances. DyCo3D~\cite{DyCo3D} improves pointgroup by introducing a dynamic-convolution-based instance decoder.  Observing that non-end-to-end clustered-based methods often exhibit over-segmentation and under-segmentation, Chen \etal ~\cite{chen2021hierarchical} and Liang \etal~\cite{liang2021instance} proposed mid-level shape representation to generate instance proposals hierarchically in an end-to-end training manner. Liu \etal~\cite{liu2020learning} approximate the distributions of centers to select center candidates for instance prediction. As mentioned in \cref{sec:intro}, most cluster-based methods treat semantic segmentation and instance feature learning as multitasks; only the works of \cite{ASIS} and \cite{zhao2020jsnet} fuse the network features of the instance prediction branch and the semantic segmentation branch to improve the performance of both branches. Unlike the pointwise fusion of \cite{ASIS} and \cite{zhao2020jsnet}, our method fuses instance features in a nonlocal manner guided by semantic outputs, which is more robust and effective.

\myparagraph{Part instance segmentation}  Different from object-level 3D instance segmentation, part-level 3D instance segmentation is less studied due to limited annotated data and the difficulty brought by geometry-similar but semantics-different shape parts. Mo \etal~\cite{Mo2019} present PartNet --- a large-scale dataset of 3D objects with fine-grained, instance-level, and hierarchical part information. For the part instance segmentation task, they developed a detection-by-segmentation method and trained a specific network to extract part instances per structural level, where the semantic hierarchy was used for part instance segmentation. Other object-level instance segmentation methods, such as \cite{DyCo3D,zhang2021}, have also been extended to the task of part instance segmentation, but they do not use the semantic hierarchy. Yu \etal~\cite{Yu2019} further enriched PartNet with information about the binary hierarchy and designed a recursive neural network to perform recursive binary decomposition to extract 3D parts. Our multi- and cross-level instance feature fusion uses semantic hierarchy to improve instance center prediction. Furthermore, the use of semantic region centers assists instance grouping. The semantic region centers serve the role of symmetric centers of a group of semantics-same part instances and provide weak supervision to the training.

\section{Methodology} \label{sec:method}
In this section, we first introduce our baseline neural network for single-level and multi-level 3D part instance segmentation in \cref{subsec:baseline}, then present the model enhanced by our semantic segmentation-assisted instance feature fusion module in \cref{subsec:fusion} and the semantic region center prediction module in \cref{subsec:semantic_center}.  

\subsection{Baseline network} \label{subsec:baseline}

\begin{figure}[t]
\centering
  \begin{overpic}[width=0.9\linewidth]{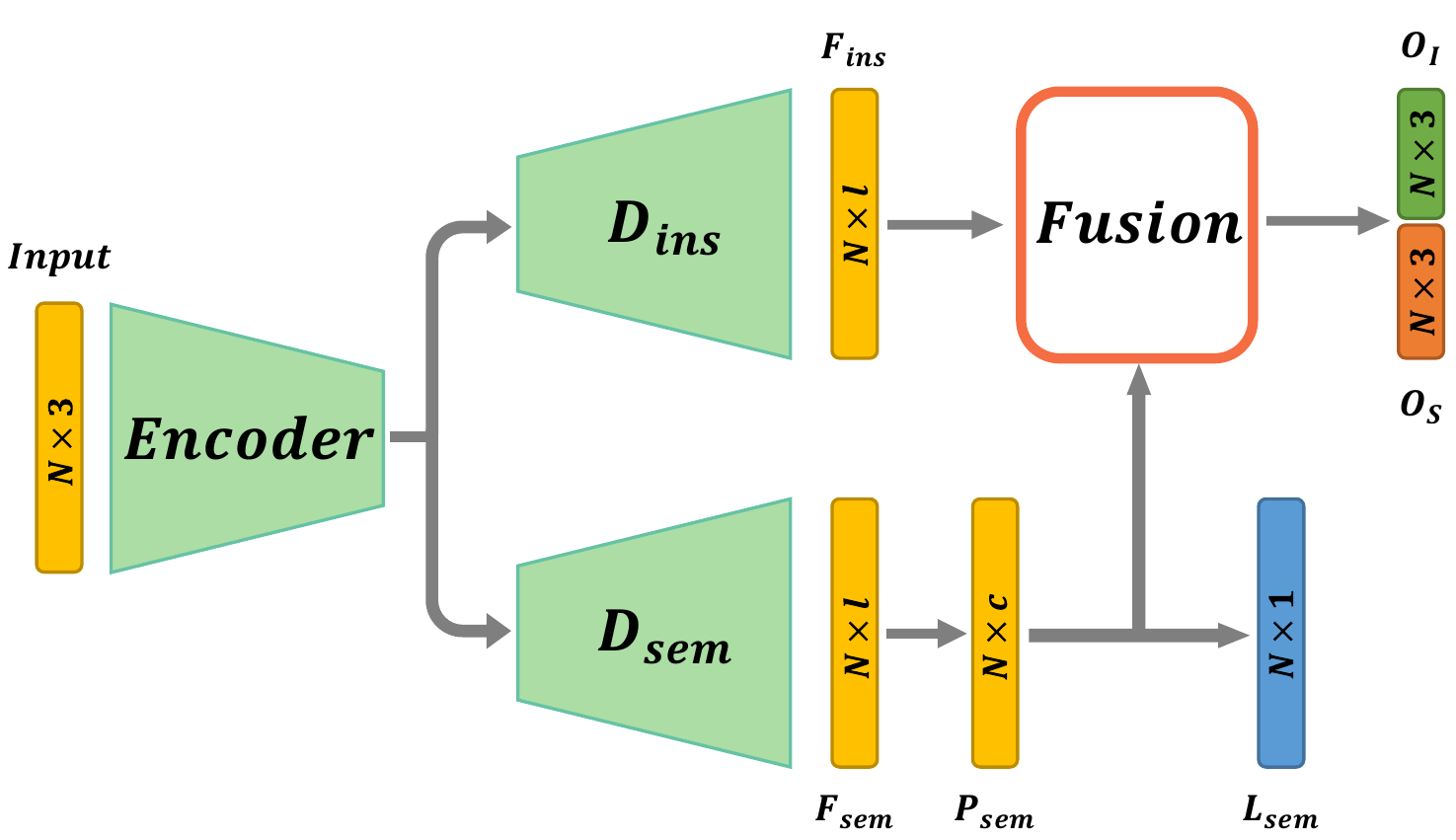}
   \end{overpic}
  \caption{
    Illustration of our network architecture for single-level part instance segmentation. The network takes a 3D point cloud as input. $N$ is the point number. A shared encoder and two parallel decoders $D_{sem}, D_{ins}$ are used to output the pointwise semantic feature $F_{sem}$ and instance feature $F_{ins}$ to predict the point semantic label $L_{sem}$ and the offset vector $O_I$ to the instance center, and the offset vector $O_S$ to the semantic region center. The feature fusion module aggregates the instance features of points according to semantic segmentation probability vectors to improve the offset prediction. 
  }
  \label{fig:network} \vspace{-5mm}
\end{figure}

Our baseline network follows the encoder-decoder paradigm. The input to the encoder is a set of 3D points $\mS$ in which each point may be equipped with additional signals such as point normal and RGB color. Two parallel decoders are concatenated after the encoder to predict the point-wise semantic labels and the point offset to its corresponding instance center, named \emph{semantic decoder} $D_{sem}$ and \emph{instance decoder} $D_{ins}$, respectively.  The baseline network is depicted in \cref{fig:network}, where the fusion module and  semantic region center will be introduced in \ref{subsec:fusion} and \ref{subsec:semantic_center}, respectively. 

The input points are shifted by the predicted offsets, and the shifted points with the same semantics are clustered into multiple 3D instances via the mean-shift algorithm~\cite{Comaniciu2002MeanSA}. In an ideal situation, all input points are shifted to their ground truth instance centers, but in practice, the accuracy of predicted offsets affects the performance of instance clustering.

\myparagraph{Network structure} We choose O-CNN-based U-Nets~\cite{Wang2017,Wang2020} as our encoder-decoder structure. The network is built on octree-based CNNs , and its memory and computational efficiency are similar to those of other sparse convolution-based neural networks~\cite{Graham2017,choy20194d}. The input point cloud is converted to an octree first, whose non-empty finest octants store the average signal of the points contained by the octants. Both $D_{sem}$ and $D_{ins}$ output point-wise features via trilinear interpolation on sparse voxels: $F_{sem}, F_{ins} \in \mathbb{R}^{N\times l}$, where $N$ is the number of points and $l$ is the dimension of feature vectors.

\begin{figure*}[t]
\centering
  \vspace{3mm}
  \begin{overpic}[width=0.8\linewidth]{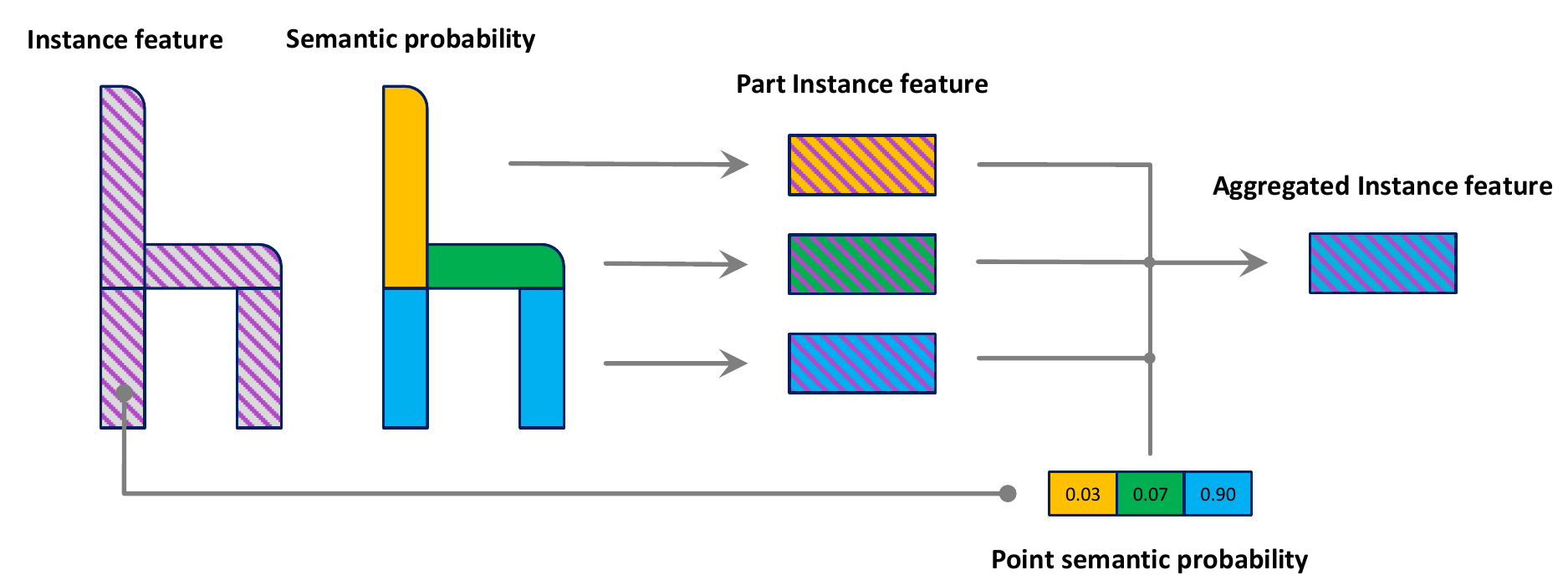}
   \end{overpic}
  \caption{
    Semantic segmentation-assisted instance feature fusion pipeline. Given the per-point instance feature and semantic probability, we get the part instance features according to the instance feature of points associated with each semantic part. Then we obtain the aggregated instance feature for each point by combining part instance features using its semantic probability.
  }
  \label{fig:instance_feature_fusion}
  \vspace{-5mm}
\end{figure*}

\myparagraph{Semantic prediction and offset prediction} A two-layer MLP  is used to convert $F_{sem}$ to the segmentation probability $P_{sem}\in \mathbb{R}^{N \times c}$,  where $c$ is the number of semantic classes.  The segmentation label $L_{sem}$ is then determined from $P_{sem}$.  The loss for training semantic segmentation is the standard cross-entropy loss. 
\begin{equation}
    L_{semantic} = \frac{1}{N}\sum_{i=1}^{N}CE(p_i, p^{*}_i).
\end{equation}
Here, $p^*$ is the semantic label.

Parallel to the semantic branch, another two-layer MLP maps $F_{ins}$ to the offset tensor $O_{I} \in \mathbb{R}^{N\times3}$, which is used to shift the input points to the center of the target instance. The loss for predicting the offsets is the $L_2$ loss between the prediction and the ground-truth offsets.
\begin{equation}
    L_{offset} = \frac{1}{N}\sum_{i=1}^{N}||o_i-o^{*}_i||_2.
\end{equation}
Here, $o^*$ is the ground-truth offset.

\myparagraph{Instance clustering} During the test phase, the network outputs pointwise semantics and offset vectors. We use the mean-shift algorithm to group the shifted points with the same semantics into disjointed instances.

\myparagraph{Multi-level part instances} For shapes with hierarchical and multi-level part instances, there are two na\"{i}ve way to extend the baseline network: (1) train the baseline network for each level individually; (2) revise the baseline network to output multi-level semantics and multi-level offset vectors simultaneously by adding multi-prediction branches after $F_{ins}$ and $F_{sem}$. We denote $K$ as the level number, add a superscript $k$ to all the symbols defined above to distinguish features at the $k$-th level, like $F_{sem}^{(k)}, F_{ins}^{(k)}, P_{sem}^{(k)}, c^{(k)}, O_I^{(k)}$.

\subsection{Semantic segmentation-assisted instance feature fusion} \label{subsec:fusion}

\subsubsection{Single-level instance feature fusion}
As the points within the same instance possess the same instance center, it is essential to aggregate the instance features over these points to regress the offset to the instance center robustly. However, these points are not known during the network inference stage and they are also the objective of the task. The semantic decoder branch can predict the semantic region composed by a set of part instances; we can aggregate the instance features over the semantic parts to provide nonlocal guidance to the input points. 
We propose a semantic segmentation-assisted instance feature fusion module that contains two steps. In the first step, for each semantic part, we compute the instance feature based on the points associated with this part. Each point is associated with an aggregated instance feature from semantic parts in the second step according to its semantic probability vector. The instance feature fusion pipeline is illustrated in \cref{fig:instance_feature_fusion}. Our feature aggregation procedure is as follows.

\myparagraph{Part instance feature}
We first aggregate the instance features with respect to the semantic label $m \in \{1,\ldots,c\}$ over the input:
\begin{equation}
    Z_m := \dfrac{\sum_{\mp \in \mS}P_{sem}(\mp)|_{m}\cdot  F_{ins}(\mp)}{\sum_{\mp \in \mS}P_{sem}(\mp)|_{m}}.
\end{equation}
$Z_m$ is the aggregated instance feature for the semantic part with semantic label of $m$, $P_{sem}(\mp)|_{m}$ is the probability value of point $\mp$ with respect to the semantic label $m$.

\myparagraph{Aggregated instance feature}
For each point $\mp$, we aggregate the instance feature $Z_m$s using the semantic probability of $\mp$ as follows:
\begin{equation} \label{eq:softfusion}
    \hat{F}(\mp) = \sum_{m=1}^c P_{sem}(\mp)|_{m} \cdot Z_m.
\end{equation}

\begin{figure*}[t]
\centering
  \begin{overpic}[width=0.9\linewidth]{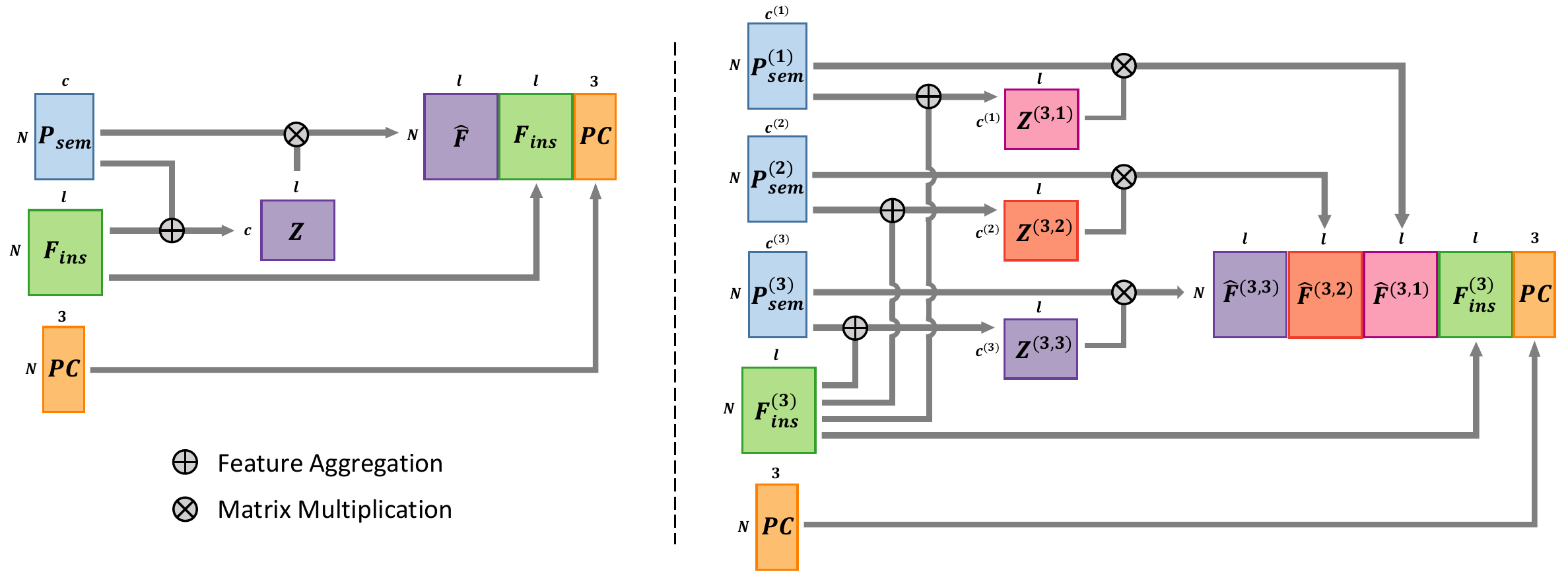}
    \put(19,0){\small (a) }
    \put(75,0){\small  (b)}
   \end{overpic}
  \caption{
    Semantic segmentation-assisted instance feature fusion for single-level and cross-level. (a) Single-level instance feature fusion. Instance features $\mathbf{F_{ins}}$ are aggregated to $\mathbf{\hat{F}}$, with the help of semantic probability vectors $\mathbf{P_{sem}}$. $\mathbf{\hat{F}}$, $\mathbf{F_{ins}}$ and the point position $\mathbf{PC}$ are assembled to form the fused instance features $\mathbf{F_{fusion}}$.  (b) Cross-level instance feature fusion for a 3-level part instance segmentation. The fused features at the 3rd level are depicted. For clarity, we omit fused features at other levels.
  }
  \label{fig:fusion_module}
  \vspace{-5mm}
\end{figure*}

The above equations for all points can be written in matrix form: $\mathbf{Z} =  \left(\mathbf{P_{sem}}  / \left(\mathbf{I_1}\mathbf{P_{sem}}  \right) \right)^T  \mathbf{F_{ins}}, \mathbf{\hat{F}} = \mathbf{P_{sem}}  \mathbf{Z}$, where $\mathbf{Z} \in \mathbb{R}^{c \times l }, \mathbf{P_{sem}} \in \mathbb{R}^{N\times c}, \mathbf{F_{ins}} \in \mathbb{R}^{N \times l}, \mathbf{\hat{F}} \in \mathbb{R}^{N \times l}$, $\mathbf{I_1} $ is an $N\times N$ matrix with all ones, and ``$/$'' represents element-wise division.

We concatenate the aggregated instance feature $\hat{F}(\mp)$, the local instance feature $F_{ins}(\mp)$ and the position of $\mp$ to form a fused instance feature $F_{fusion}(\mp) :=[\hat{F}(\mp), F_{ins}(\mp), \mp]$, and use it to predict the instance center offset. \cref{fig:fusion_module}-(a) illustrates our feature fusion module for a single level.  The overall network structure is shown in \cref{fig:network}. 

\subsubsection{Multi-level instance feature fusion} For shapes with multi-level part instances, our single-level instance feature fusion can be applied to each level individually. The na\"{i}vely extended baseline networks (\cref{subsec:baseline}) can benefit from this kind of instance feature fusion for multi-level part instance segmentation. 

\subsubsection{Cross-level instance feature fusion} When multi-level part instances and semantic segmentation exhibit a hierarchical relationship, \ie the fine-level part instances are contained within the coarser-level part instances and can inherit the semantics from their parent level, we leverage the semantic segmentation in multi-levels to fuse instance features at each level, we call our strategy \emph{cross-level instance feature fusion}. The exact fusion procedure is as follows. 

\myparagraph{Instance feature aggregation}
On level $k$, we aggregate the instance features using semantic probability vectors at the $r$-th level:
\begin{equation}
    Z_m^{(k,r)} := \dfrac{\sum_{\mq \in \mS}P_{sem}^{(r)}(\mq)|_{m}\cdot  F^{(k)}_{ins}(\mq)}{\sum_{\mq \in \mS}P^{(r)}_{sem}(\mq)|_{m}}, \; m \in \{1,\ldots, c^{(r)}\}.
\end{equation}
$Z_m^{(k,r)}$s are then averaged at point $\mp$ at the $k$-th level:
\begin{equation} 
    \hat{F}^{(k,r)}(\mp) = \sum_{m=1}^{c^{(r)}} P^{(r)}_{sem}(\mp)|_{m} \cdot Z^{(k,r)}_m.
\end{equation}
The fused instance feature of $\mp$ at the $k$-th level is defined as follows:
\begin{equation*} 
    F_{fusion}^{(k)}(\mp) :=[\hat{F}^{(k,1)}(\mp), \cdots, \hat{F}^{(k,K)}(\mp),  F^{(k)}_{ins}(\mp), \mp].
\end{equation*}
It is mapped to offset vectors at the $k$-th level by an MLP layer. We illustrate the cross-level instance feature fusion in \cref{fig:fusion_module}-(b).

\subsection{Semantic region center} \label{subsec:semantic_center}

\begin{figure}[t]
\centering
  \begin{overpic}[width=0.85\linewidth]{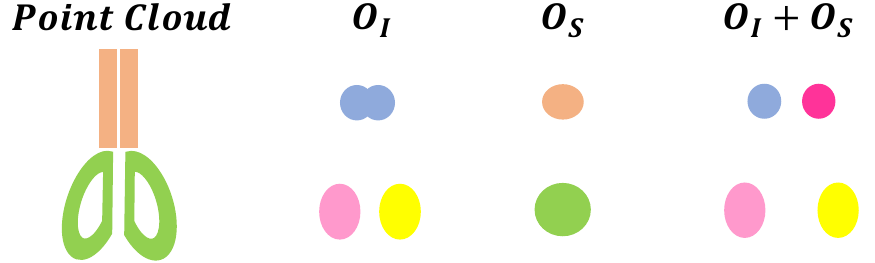}
    \put(11.2,-3){\small (a)}
    \put(40,-3){\small (b)}
    \put(62,-3){\small (c)}
    \put(88,-3){\small (d)}
   \end{overpic}
  \caption{
    Illustration of the use of semantic region centers. (a) Input point cloud of a scissor shape. Ground-truth part instances are colored according to their semantics. (b) Predicted instance centers. (c) Predicted semantic region centers. (d) By pushing the predicted instance centers away from the predicted semantic region centers, the shifted instance centers of the scissor blades become more distinguishable than in(b).
  }
  \label{fig:semantic_instance_center} \vspace{-5mm}
\end{figure}
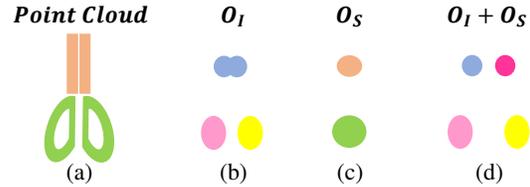

During the test phase, we use the mean-shift algorithm to split the offset-shifted points with the same semantics into different instances. For 3D instances which are close to each other, like two blades of a scissor shown in 
\cref{fig:semantic_instance_center}-(a), it is difficult to separate the points belonging to them using mean-shift or other 3D point clustering algorithms, as the instance centers are very close to each other (see \cref{fig:semantic_instance_center}-(b)). We introduce the concept of semantic region center, which is the center of semantically -same instance centers. The semantic region center is usually the center of symmetrically arranged parts for human-made shapes \cref{fig:semantic_instance_center}-(c) illustrates the semantic region centers. To make instance clustering easy, the instance centers can be further shifted away from the semantic region center, as shown \cref{fig:semantic_instance_center}-(d). In the offset prediction branch of our network, we also add the offset prediction $O_S$ to the center of the semantic region for each point.

In the instance clustering step, we shift the input points as follows:
\begin{equation} \label{eq:shiftcenter}
    \hat{\mp} :=\mp + O_I(\mp) + \lambda \cdot \dfrac{O_I(\mp) - O_S(\mp)}{||O_I(\mp) - O_S(\mp)||}.
\end{equation}
Here $\mp \in \mS$, $\lambda > 0$.

\begin{table*}[t]

    \resizebox{\textwidth}{!}{%

        \begin{tabular}{@{}c|c|c|*{24}{c}@{}}
            \toprule
             &  Level   & Avg           & \rotatebox[origin=lB]{60}{Bag} & \rotatebox[origin=lB]{60}{Bed} & \rotatebox[origin=lB]{60}{Bottle} & \rotatebox[origin=lB]{60}{Bowl} & \rotatebox[origin=lB]{60}{Chair} & \rotatebox[origin=lB]{60}{Clock} & \rotatebox[origin=lB]{60}{Dish} & \rotatebox[origin=lB]{60}{Disp} & \rotatebox[origin=lB]{60}{Door} & \rotatebox[origin=lB]{60}{Ear} & \rotatebox[origin=lB]{60}{Faucet} & \rotatebox[origin=lB]{60}{Hat} & \rotatebox[origin=lB]{60}{Key} & \rotatebox[origin=lB]{60}{Knife} & \rotatebox[origin=lB]{60}{Lamp} & \rotatebox[origin=lB]{60}{Laptop} & \rotatebox[origin=lB]{60}{Micro} & \rotatebox[origin=lB]{60}{Mug} & \rotatebox[origin=lB]{60}{Fridge} & \rotatebox[origin=lB]{60}{Scis} & \rotatebox[origin=lB]{60}{Stora} & \rotatebox[origin=lB]{60}{Table} & \rotatebox[origin=lB]{60}{Trash} & \rotatebox[origin=lB]{60}{Vase} \\ \toprule
            \multirow{4}{*}{\rotatebox[origin=c]{90}{SGPN}}
             & Coarse   & 55.7          & 38.8                           & 29.8                           & 61.9                              & 56.9                            & 72.4                             & 20.3                             & 72.2                            & 89.3                            & 49.0                            & 57.8                           & 63.2                              & 68.7                           & 20.0                           & 63.2                             & 32.7                            & \textbf{100.0}                    & 50.6                             & 82.2                           & 50.6                              & 71.7                            & 32.9                             & 49.2                             & 56.8                             & 46.6                            \\
             & Middle   & 29.7          & -                              & 15.4                           & -                                 & -                               & 25.4                             & -                                & 58.1                            & -                               & 25.4                            & -                              & -                                 & -                              & -                              & -                                & 21.7                            & -                                 & 49.4                             & -                              & 22.1                              & -                               & 30.5                             & 18.9                             & -                                & -                               \\
             & Fine   & 29.5          & -                              & 11.8                           & 45.1                              & -                               & 19.4                             & 18.2                             & 38.3                            & 78.8                            & 15.4                            & 35.9                           & 37.8                              & -                              & -                              & 38.3                             & 14.4                            & -                                 & 32.7                             & -                              & 18.2                              & -                               & 21.5                             & 14.6                             & 24.9                             & 36.5                            \\ \cmidrule{2-27}
             & Avg & 46.8          & 38.8                           & 19.0                           & 53.5                              & 56.9                            & 39.1                             & 19.3                             & 56.2                            & 84.1                            & 29.9                            & 46.9                           & 50.5                              & 68.7                           & 20.0                           & 50.8                             & 22.9                            & \textbf{100.0}                    & 44.2                             & 82.2                           & 30.3                              & 71.7                            & 28.3                             & 27.6                             & 40.9                             & 41.6                            \\ \toprule
            \multirow{4}{*}{\rotatebox[origin=c]{90}{PartNet}}
             & Coarse   & 62.6          & 64.7                           & 48.4                           & 63.6                              & 59.7                            & 74.4                             & 42.8                             & 76.3                            & 93.3                            & 52.9                            & 57.7                           & 69.6                              & 70.9                           & 43.9                           & 58.4                             & 37.2                            & \textbf{100.0}                    & 50.0                             & 86.0                           & 50.0                              & 80.9                            & 45.2                             & 54.2                            & 71.7                             & 49.8                            \\
             & Middle   & 37.4          & -                              & 23.0                           & -                                 & -                               & 35.5                             & -                                & 62.8                            & -                               & 39.7                            & -                              & -                                 & -                              & -                              & -                                & 26.9                            & -                                 & 47.8                             & -                              & 35.2                              & -                               & 35.0                             & 31.0                             & -                                & -                               \\
             & Fine   & 36.6          & -                              & 15.0                           & 48.6                              & -                               & 29.0                             & 32.3                             & 53.3                            & 80.1                            & 17.2                            & 39.4                           & 44.7                              & -                              & -                              & 45.8                             & 18.7                            & -                                 & 34.8                             & -                              & 26.5                              & -                               & 27.5                             & 23.9                             & 33.7                             & 52.0                            \\ \cmidrule{2-27}
             & Avg & 54.4          & 64.7                           & 28.8                           & 56.1                              & 59.7                            & 46.3                             & 37.6                             & 64.1                            & 86.7                            & 36.6                            & 48.6                           & 57.2                              & 70.9                           & 43.9                           & 52.1                             & 27.6                            & \textbf{100.0}                    & 44.2                             & 86.0                           & 37.2                              & 80.9                            & 35.9                             & 36.4                             & 52.7                             & 50.9                            \\ \toprule
            \multirow{4}{*}{\rotatebox[origin=c]{90}{PE}}
             & Coarse   & 65.1          & 64.6                           & 51.4                           & 63.1                              & 72.0                            & 77.1                             & 41.1                             & 76.9                            & 95.3                            & 61.2                            & 66.5                           & 73.1                              & 71.8                           & \textbf{48.6}                  & \textbf{76.5}                    & 37.1                            & \textbf{100.0}                    & 50.5                             & 90.9                           & 50.5                              & \textbf{88.6}                   & 47.3                             & 40.3                             & 69.0                             & 48.7                            \\
             & Middle   & 40.4          & -                              & 31.0                           & -                                 & -                               & 38.6                             & -                                & 64.2                            & -                               & 36.9                            & -                              & -                                 & -                              & -                              & -                                & 31.0                            & -                                 & 51.2                             & -                              & 37.3                              & -                               & 42.0                             & 31.5                             & -                                & -                               \\
             & Fine   & 39.8          & -                              & 26.2                           & 50.7                              & -                               & 34.7                             & 30.2                             & 50.0                            & 82.0                            & 25.7                            & 43.2                           & 55.6                              & -                              & -                              & 44.4                             & 20.3                            & -                                 & 37.0                             & -                              & 31.1                              & -                               & 34.2                             & 25.5                             & 37.7                             & 47.6                            \\ \cmidrule{2-27}
             & Avg & 57.5          & 64.6                           & 36.2                           & 56.9                              & 72.0                            & 50.1                             & 35.6                             & 63.7                            & 88.7                            & 41.3                            & 54.9                           & 64.4                              & 71.8                           & \textbf{48.6}                  & \textbf{60.5}                    & 29.5                            & \textbf{100.0}                    & 46.2                             & 90.9                           & 39.6                              & \textbf{88.6}                   & 41.2                             & 32.4                             & 53.4                             & 48.1                            \\ \toprule
            \multirow{4}{*}{\rotatebox[origin=c]{90}{\textbf{Ours}}}
             & Coarse   & \textbf{71.2} & \textbf{80.5}                  & \textbf{54.1}                  & \textbf{66.9}                     & \textbf{84.2}                   & \textbf{84.1}                    & \textbf{51.2}                    & \textbf{79.9}                   & \textbf{97.2}                   & \textbf{76.8}                   & \textbf{71.6}                  & \textbf{79.2}                     & \textbf{77.3}                  & 47.0                           & 67.8                             & \textbf{38.2}                            & \textbf{100.0}                    & \textbf{62.5}                    & \textbf{91.8}                  & \textbf{57.4}                     & 86.8                            & \textbf{56.4}                    & \textbf{65.3}                             & \textbf{79.7}                    & \textbf{53.8}                   \\
             & Middle   & \textbf{49.7} & -                              & \textbf{45.5}                  & -                                 & -                               & \textbf{45.7}                    & -                                & \textbf{73.2}                   & -                               & \textbf{52.0}                   & -                              & -                                 & -                              & -                              & -                                & \textbf{30.9}                   & -                                 & \textbf{62.5}                    & -                              & \textbf{48.2}                     & -                               & \textbf{53.3}                    & \textbf{36.2}                    & -                                & -                               \\
             & Fine   & \textbf{47.8} & -                              & \textbf{40.9}                  & \textbf{55.9}                     & -                               & \textbf{38.2}                    & \textbf{37.1}                    & \textbf{56.5}                   & \textbf{87.4}                   & \textbf{41.3}                   & \textbf{53.7}                  & \textbf{59.1}                     & -                              & -                              & \textbf{48.8}                    & \textbf{21.7}                   & -                                 & \textbf{49.7}                    & -                              & \textbf{44.1}                     & -                               & \textbf{44.0}                    & \textbf{28.9}                    & \textbf{51.3}                    & \textbf{54.6}                   \\ \cmidrule{2-27}
             & Avg & \textbf{64.1} & \textbf{80.5}                  & \textbf{46.8}                  & \textbf{61.4}                     & \textbf{84.2}                   & \textbf{56.0}                    & \textbf{44.2}                    & \textbf{69.9}                   & \textbf{92.3}                   & \textbf{56.7}                   & \textbf{62.7}                  & \textbf{69.2}                     & \textbf{77.3}                  & 47.0                           & 58.3                             & \textbf{30.3}                   & \textbf{100.0}                    & \textbf{58.2}                    & \textbf{91.8}                  & \textbf{49.9}                     & 86.8                            & \textbf{51.2}                    & \textbf{43.5}                    & \textbf{65.5}                    & \textbf{54.2}                   \\

            \bottomrule
        \end{tabular}
    }
    \caption{Part instance segmentation results of the test set on PartNet~\cite{Mo2019}. We report part-category $AP_{50}$ on three instance levels. The results of other methods are reported by PE~\cite{zhang2021}. Bold numbers are better. Some shape categories, masked by dashed lines, have no middle- and fine-level instances for benchmark.
    }
    \label{tab:partnet-map50} \vspace{-3mm}
\end{table*}

\section{Experiments and Analysis} \label{sec:result}
We design a series of experiments and ablation studies to demonstrate the efficacy of our approach and its superiority to other fusion schemes, including multi-level part instance segmentation on PartNet~\cite{Mo2019} (\cref{subsec:partnet}), and instance segmentation on indoor scene datasets (\cref{subsec:indoor}): ScanNet~\cite{dai2017scannet} and S3DIS~\cite{S3DIS}. 

\subsection{Part instance segmentation on PartNet} \label{subsec:partnet}

\subsubsection{Experiments and comparison}

\myparagraph{Dataset}
PartNet is a large-scale dataset with fine-grained and hierarchical part annotations. It contains more than 570k part instances over 26,671 3D models covering 24 object categories. It provides coarse-, middle-, and fine-grained part instance annotations. 

\myparagraph{Network configuration}
The encoder and decoders of our O-CNN-based U-Net had five levels of domain resolution, and the maximum depth of the octree was six. The dimension of the feature was set to \num{64}. Details of the U-Net structure are provided in Appendix A. We implemented our network in the TensorFlow framework~\cite{tensorflow2015-whitepaper}. The network was trained with $100000$ iterations with a batch size of 8. We used the SGD optimizer with a learning rate of \num{0.1} and decay two times with the factor of \num{0.1} at the 50000-th and 75000-th iterations. Our code and trained models are available.

\myparagraph{Data processing}
The input point cloud contained \num{10000} points and was scaled into a unit sphere. During training, we also augmented each shape by a uniform scaling with the scale ratio of $[0.75, 1,25]$, a random rotation whose pitch, yaw, and roll rotation angles were less than \ang{10}, and random translations along each coordinate axis within the interval $[-0.125, 0.125]$.
The train/test split is provided in PartNet. Note that not all categories have three-level part annotations. During training, we duplicated the labels at the coarser level to the finer level, if the latter was missing, to mimic the three-level shape structure. During the test phase, we only evaluated the output from the levels which exist in the data. The ground-truth instance centers and semantic region centers were pre-computed according to the semantic labels and part instances of PartNet.

\myparagraph{Experiment setup}
We set $\lambda = 0.05$ for \cref{eq:shiftcenter}. We used the mean-shift implementation implemented in scikit-learn~\cite{scikit-learn}. The default bandwidth of mean-shift was set to \num{0.1}. All our experiments were conducted on an Azure Linux server with Intel Xeon Platinum 8168 CPU (\SI{2.7}{GHz}) and Tesla V100 GPU (\SI{16}{GB} memory). Our baseline network with cross-level fusion was the default configuration. In practice, we found that stopping the gradient from the fusion module to the semantic decoder helps maintain the semantic segmentation accuracy and slightly improves the instance segmentation. So, we enabled gradient stopping by default. An ablation study on gradient stopping is provided in \cref{subsub:ablation}.

\myparagraph{Evaluation metrics} We used \emph{per-category mAP score} with the IoU  threshold of 0.25, 0.5 and 0.75 to evaluate the quality of part instance segmentation. They are denoted by $AP_{25}$, $AP_{50}$ and $AP_{75}$. s-$AP_{50}$ is the metric proposed by ~\cite{Mo2019}, which averages the precision over the shapes.

\myparagraph{Performance report and comparison}
We report $AP_{50}$ of our approach in all 24 shape categories in \cref{tab:partnet-map50}. We also report the performance of three comparison approaches: SGPN~\cite{SGPN}, PartNet~\cite{Mo2019}, and PE~\cite{zhang2021}. The results are averaged over three levels of granularity. Our method outperformed the best competitor PE~\cite{zhang2021} by \SI{6.6}{\percent}, and also achieved the best performance in most categories. Our approach was also the best on other evaluation metrics, as shown in \cref{tab:partnet-all-metric}. Appendix C reports the per-category results of $AP_{25}$, $AP_{75}$ and s-$AP_{50}$. As DyCo3D~\cite{DyCo3D} only performed instance segmentation experiments in four categories of the PartNet dataset, we compare it with our approach on these categories separately in \cref{tab:partnet-four-category}. Our method outperformed DyCo3D by a large margin.

\begin{table}[t]
    \centering
    \resizebox{0.9\columnwidth}{!}{
        \begin{tabular}{l|*{18}{c}}
            \toprule
                                       & $AP_{25}$ & $AP_{50}$ & $AP_{75}$ & s-$AP_{50}$ & mIoU\\\midrule
            SGPN~\cite{SGPN}           & -            & 46.8      & -            & 64.2      & -   \\
            PartNet~\cite{Mo2019}      & 62.8         & 54.4      & 38.9         & 72.2     & -    \\
            PE~\cite{zhang2021}        & 66.5         & 57.5      & 41.7         & -        & -    \\
            Ours                       & \textbf{72.1}         & \textbf{64.1}      & \textbf{49.7}         &\textbf{76.1}  & 66.1      \\
            \bottomrule
        \end{tabular}}
    \caption{Part instance segmentation on the test set of PartNet. $AP_{25}, AP_{50}, AP_{75}$, s-$AP_{50}$ are averaged over three levels. The results of other methods are reported by PartNet~\cite{Mo2019} and PE~\cite{zhang2021}.
    }
    \label{tab:partnet-all-metric}
\end{table}

\begin{table}[t]
    \centering
    \resizebox{0.75\columnwidth}{!}{
        \begin{tabular}{l|c|*{4}{c}} 
            \toprule
                            & Level           & Chair & Lamp & Stora. & Table \\\midrule
            \multirow{4}{*}{\rotatebox[origin=c]{90}{DyCo3D}}           & Coarse            & 81.0      & 37.3            & 44.5 & 55.0     \\
            & Middle            & 41.3      & 28.8            & 38.9 & 32.5     \\
            & Fine            & 33.4      & 20.5            & 30.4  & 24.9   \\\cmidrule{2-6}
            & Avg            & 51.9      & 28.9            & 37.9  & 37.5   \\\midrule
            \multirow{4}{*}{\rotatebox[origin=c]{90}{Ours}}         & Coarse            & \textbf{84.1}      & \textbf{38.2}            & \textbf{56.4} & \textbf{65.3}     \\
            & Middle            & \textbf{45.7}      & \textbf{30.9}            & \textbf{53.3} & \textbf{36.2}     \\
            & Fine            & \textbf{38.2}      & \textbf{21.7}            & \textbf{44.0} & \textbf{28.9}     \\\cmidrule{2-6}
            & Avg            & \textbf{56.0}      & \textbf{30.3}            & \textbf{51.2} & \textbf{43.5}     \\
            \bottomrule
        \end{tabular}}
    \caption{Part instance segmentation on the four categories of PartNet. $AP_{50}$ is reported.}
    \label{tab:partnet-four-category} \vspace{-5mm}
\end{table}

\subsubsection{Ablation study} \label{subsub:ablation}
We validated our network design on PartNet instance segmentation, especially for the fusion module and the semantic region centers. The variants of our network are listed below.

\begin{enumerate}[leftmargin=10pt,nosep]
  \item[-] \textbf{Single-level baseline}: the network trained for each level individually without using the fusion module. 
  \item[-] \textbf{Multi-level baseline}: the network trained for multi-levels simultaneously without using the fusion module. 
  \item[-] \textbf{Single-level fusion}: Single-level baseline with single-level fusion. 
  \item[-] \textbf{Multi-level fusion}: Multi-level baseline with single-level fusion on each level. 
  \item[-] \textbf{Cross-level fusion}: Multi-level baseline with cross-level fusion.
 \end{enumerate}
 For each variant, we use symbol $\dagger$ to indicate that the predicted semantic region centers are not used for instance clustering. The optimal variant is \emph{cross-level fusion}. The performance of each variant is reported in \cref{tab:partnet-ablation}.

\myparagraph{Single-level baseline versus multi-level baseline} The performances of \textit{single-level baseline} and \textit{multi-level baseline} in the same setting (w. or w/o fusion and semantic region center) are not much different. However, the training effort of \textit{multi-level baseline} is much lower. There are a total of 50 levels for all 24 categories of PartNet. The \textit{single-level baseline} must train 50 networks, while the \textit{multi-level baseline} only needs to train 24 networks.

\myparagraph{Fusion module} It is clear that the performance of all baselines with the fusion modules improved.  \textit{Single-level fusion} and \textit{multi-level fusion} increase $AP_{50}$ by +$3.9$ and +$4.4$ points compared to their baselines. 
\textit{Cross-level fusion} surpasses them at $AP_{50}$ by +$2.0$ and +$1.6$ points. Here, the network of cross-level fusion has a slightly large network size. On Chair category, the network parameters of cross-level fusion, multi-level fusion, multi-level baselines are \SI{8.13}{M}, \SI{7.98}{M}, and \SI{7.89}{M}, respectively.

\begin{table}[t]
    \centering
    \resizebox{0.85\columnwidth}{!}{
        \begin{tabular}{l|*{5}{c}}
            \toprule
                                                   & $AP_{25}$ & $AP_{50}$ & $AP_{75}$ & s-$AP_{50}$ & mIoU \\\midrule
            single-level baseline$^{\dagger}$      & 67.3          & 57.9      & 45.3         & 74.4  & 64.9       \\
            single-level baseline      & 67.4         & 58.2      & 45.5         & 75.0     & 64.9    \\\midrule
            single-level fusion$^{\dagger}$        & 70.4          & 61.2      & 48.8         & 74.8  & 65.4       \\
            single-level fusion       & 71.1         & 62.1      & 49.0         & 75.8      & 65.4   \\\midrule
            multi-level baseline$^{\dagger}$       & 67.1          & 57.9      & 45.0         & 74.1    & 65.0     \\
            multi-level baseline       & 67.3         & 58.1      & 45.1         & 74.7    & 65.0         \\\midrule
            multi-level fusion$^{\dagger}$   & 70.9          & 61.8      & 48.8         & 74.8      & 65.5       \\
            multi-level fusion   & 71.5         & 62.5      & 49.2         & 75.6       & 65.5      \\\midrule
            cross-level fusion$^{\dagger}$   & 71.3          & 63.1      & 48.6         & 75.2          & \textbf{66.1}   \\
            cross-level fusion   & \textbf{72.1}         & \textbf{64.1}      & \textbf{49.7}         & \textbf{76.1}       & \textbf{66.1}      \\\midrule
            cross-level fusion(gradient)$^{\dagger}$   & 71.1          & 62.2      & 48.4         & 75.0       & 65.2      \\
            cross-level fusion(gradient)   & 71.8         & 63.3      & 49.3         & 75.9        & 65.2     \\\midrule
            cross-level fusion(one-hot)$^{\dagger}$         & 70.7          & 62.4      & 48.1         & 75.0        & 65.8     \\
            cross-level fusion(one-hot)                     & 71.6          & 63.5      & 49.0         & 75.8        & 65.8   \\\midrule
            cross-level fusion(backbone)$^{\dagger}$         & 69.6          & 61.6      & 46.0         & 74.7        & 65.3     \\
            cross-level fusion(backbone)                     & 70.2          & 62.4      & 47.1         & 75.3        & 65.3   \\\midrule
            cross-level fusion(two-dir)$^{\dagger}$                       & 71.0          & 62.6      & 48.3         & 75.2        & 65.7     \\
            cross-level fusion(two-dir)                     & 71.8          & 63.6      & 48.7         & 76.0          & 65.7   \\\midrule
            ASIS fusion$^{\dagger}$                       & 68.2          & 59.0      & 45.0         & 74.7         & 65.1    \\
            ASIS fusion                     & 68.6         & 59.1      & 45.9         & 75.0        & 65.1     \\\midrule
            JSNet fusion$^{\dagger}$                       & 68.5          & 59.2      & 46.3         & 75.4        & 65.4     \\
            JSNet fusion                     & 68.8          & 59.3      & 46.6         & 75.6          & 65.4   \\
            \bottomrule
        \end{tabular}}
    \caption{Ablation studies of our approach on PartNet test data. Methods marked with $\dagger$ use the predicted instance centers only. Our default and optimal network setting is \emph{cross-level fusion}.}
    \label{tab:partnet-ablation} \vspace{-4mm}
\end{table}

\begin{figure*}[t]
\centering
  \begin{overpic}[width=0.9\linewidth]{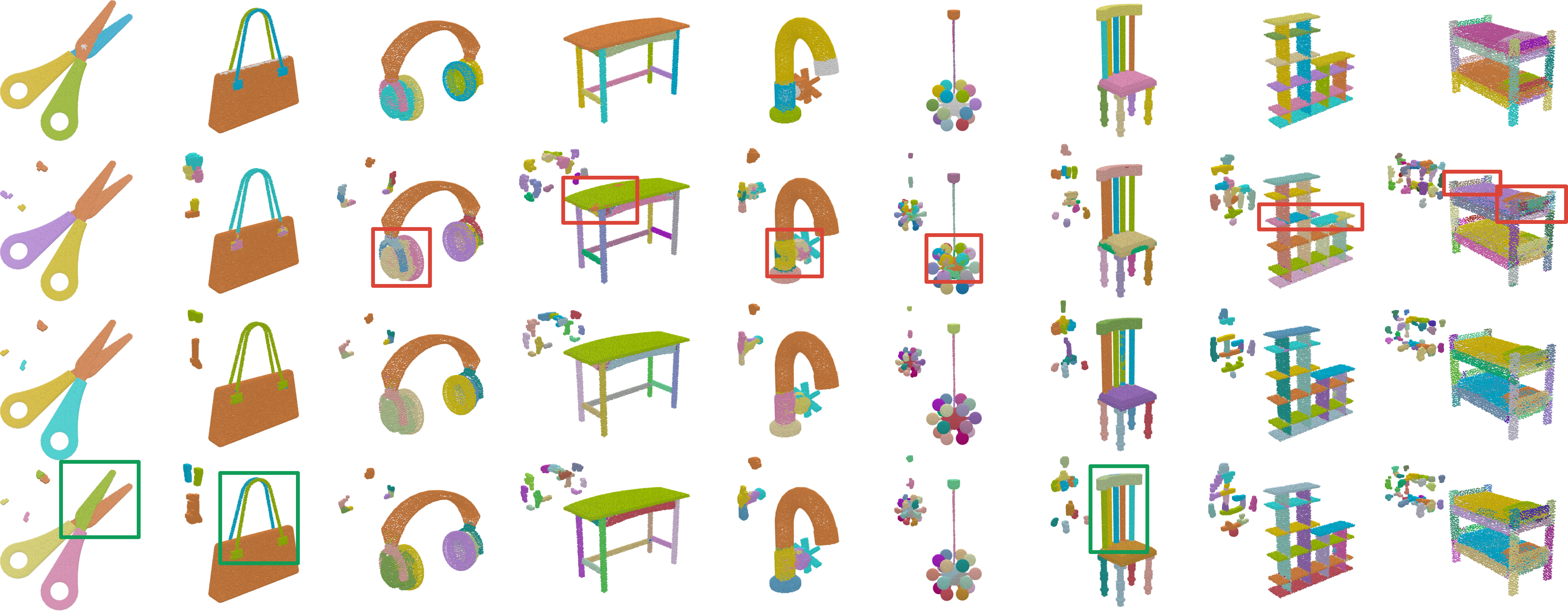}
   \end{overpic}
  \caption{Visual comparison of part instance segmentation on the test set of PartNet. Part instances at the fine level are colored with random colors. \textbf{1st row}: part instance ground-truth. \textbf{2nd row}: results of our \emph{multi-level baseline$^{\dagger}$}. \textbf{3rd row}: results of our \emph{cross-level fusion$^{\dagger}$} without using semantic region centers. \textbf{4th row}: results of our \emph{cross-level fusion} using semantic region centers. The corresponding shifted points are rendered at the top left of each instance segmentation image. Green and red boxes represent good and bad instances, respectively.
  }
  \label{fig:comparison} \vspace{-2mm}
\end{figure*}

\myparagraph{Use of semantic region centers} The instance segmentation performance is consistently improved by using semantic region centers. The improvement is also more noticeable when the fusion module is enabled to improve both the instance center prediction and the semantic region center prediction. For example, there is only +($0.2\sim0.3$) improvement when using semantic region centers on \textit{single-level baseline} and \textit{multi-level baseline}, while the improvement over \textit{cross-level fusion}$^\dagger$ is +$1.0$. 

In \cref{fig:comparison}, we present the instance segmentation results of \emph{multi-level baseline}, \emph{cross-level fusion$\dagger$} and \emph{cross-level fusion}. The predicted instance centers are more compact and distinguishable when using the fusion module. The use of semantic region centers helps further separate close instances, \eg the scissor blades in the 1st column, the bag handles in the 2nd column and the chair back frames in the 7th column.   

\myparagraph{Stopping gradient} One of the inputs of the fusion module is the semantic segmentation probability. The gradients of the fusion module can backpropagate the errors to the semantic branch. In our experiments, we found that gradient backpropagation impairs semantic segmentation and leads to slightly worse instance segmentation results (see \textit{cross-level fusion(gradient)} in \cref{tab:partnet-ablation}).

\myparagraph{Instance feature aggregation} In our instance feature fusion module, we used the semantic probability of the point to aggregate the instance features from different semantic parts. An alternative way is to aggregate the instance features of the part which the point belongs to, \ie using the one-hot version of semantic probability for each point. We found that our default fusion is better than this alternative (\textit{cross-level fusion(one-hot)} in \cref{tab:partnet-ablation}) because the instance features from different semantic parts can bring more contextual information, especially for points with fuzzy semantic probability.

\myparagraph{Network backbone} The O-CNN~\cite{Wang2017,Wang2020} backbone used in our network is different from the PointNet++~\cite{qi2017pointnetplusplus} backbone used in ~\cite{Mo2019,zhang2021}. Therefore, we also replaced the O-CNN backbone with PointNet++ for a fair comparison. As shown in \textit{cross-level fusion(backbone)} in \cref{tab:partnet-ablation}, the performance of the PointNet++ backbone with our fusion scheme is lower than that of the O-CNN backbone by $1.7$ points in $AP_{50}$, but it is still much better than \cite{Mo2019} and \cite{zhang2021}, by $+8.0$ and $+4.9$ points, respectively, in $AP_{50}$. This experiment further validates the efficacy of our approach.

\begin{figure*}[t]
\centering
  \begin{overpic}[width=0.95\linewidth]{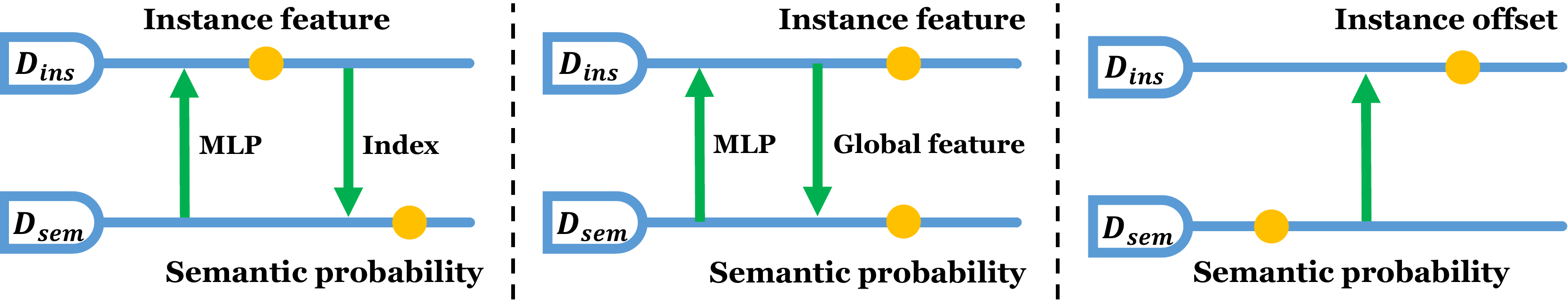}
    \put(11,-1){\small (a) ASIS }
    \put(47,-1){\small (b) JSNet}
    \put(82,-1){\small (c) Ours}
   \end{overpic}
  \caption{
    Concept illustration of the fusion schemes of ASIS~\cite{ASIS}, JSNet~\cite{zhao2020jsnet} and our method. (a) ASIS has two fusion directions. It maps the semantic feature to the instance feature using an MLP layer and uses the nearest neighbors in the instance feature space to aggregate the semantic features. (b) JSNet also has two fusion directions. It maps the semantic feature to the instance feature using an MLP layer and adds the global instance feature to the pointwise semantic feature. (c) Our fusion module has only one fusion direction: the semantic probability directly helps the aggregation of instance features in a nonlocal manner.
  }
  \label{fig:jsnet_concept_comparison} \vspace{-5mm}
\end{figure*}

\begin{figure}[t]
\centering
  \begin{overpic}[width=0.9\linewidth]{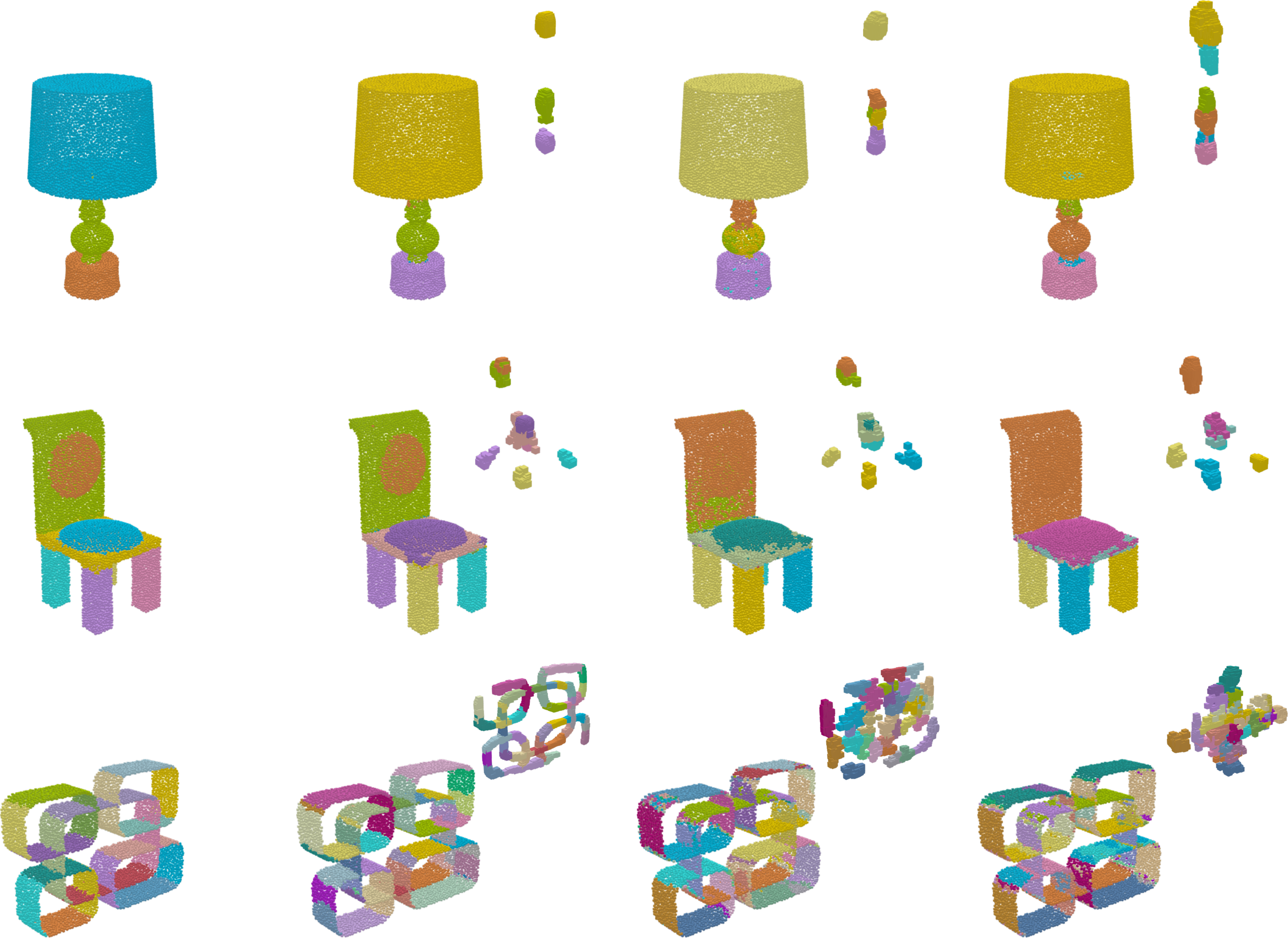}
    \put(6,-4.5){\small (a)}
    \put(29,-4.5){\small (b)}
    \put(56,-4.5){\small (c)}
    \put(81,-4.5){\small (d)}
   \end{overpic}
   \vspace{2mm}
  \caption{
    Visualization comparison of different fusion methods on PartNet. (a) Part instance ground-truth. (b) Results of our fusion module. (c) Results of ASIS fusion module. (d) Results of JSNet fusion module. The corresponding shifted points are rendered at the top right of each instance segmentation image.
  }\vspace{-2mm}
  \label{fig:fusion_comparison}
\end{figure}
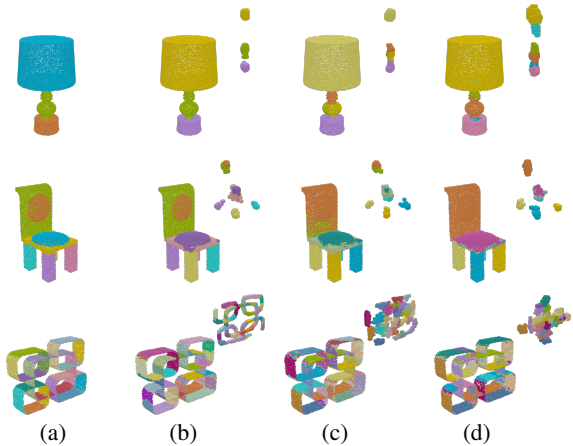

\myparagraph{Fusion scheme of ASIS~\cite{ASIS} and JSNet~\cite{zhao2020jsnet}}  We compare our fusion module with other fusion schemes proposed in ASIS~\cite{ASIS} and JSNet~\cite{zhao2020jsnet}. ASIS jointly fuses the features between the segmentation and instance branches to improve the performance, as shown in \cref{fig:jsnet_concept_comparison}-(a). It has two fusion directions: one of them maps the semantic feature to the instance feature space using an MLP layer; the other one uses K-nearest neighbors in the instance feature space to aggregate the semantic feature. Similar to ASIS, JSNet also has two fusion directions as shown in \cref{fig:jsnet_concept_comparison}-(b). One maps the semantic feature to the instance feature space, and the other adds the global instance feature to the semantic feature. Our fusion scheme differs in two aspects compared to the ASIS and JSNet fusion modules. Firstly, our fusion module has only one fusion direction, as shown in \cref{fig:jsnet_concept_comparison}-(c), which uses semantic probability to guide the instance feature aggregation. Secondly, our fusion module uses the network output of semantic branch - \textit{semantic probability} to guide the fusion of instance features, while ASIS and JSNet use the intermediate network information to fuse features. The fusion modules of ASIS and JSNet are more like enhancing the two decoders of the network, while our fusion module has a more specific target - to improve the accuracy of the predicted instance offsets. To prove the superiority of our fusion scheme, we replace our fusion module with the ASIS fusion and the JSNet fusion and integrate them with our \textit{single-level baseline} and our loss functions. We observed +$0.9$ and +$1.1$ points improvement of $AP_{50}$ over the baselines using semantic region centers (see ASIS fusion and JSNet fusion in \cref{tab:partnet-ablation}). However, the improvements are minor compared to our \textit{single-level fusion} which has +$3.9$ points improvement. In \cref{fig:fusion_comparison}, we illustrate some results generated by different fusion methods. The shifted points of our fusion module are more compact and accurate, resulting in a more reasonable segmentation of the part instances.
We also insert the other direction fusion into our fusion module by mapping the semantic feature to the instance feature space using an MLP layer. The performance is slightly worse than  \textit{cross-level fusion} due to the worse semantic segmentation results, as shown in \textit{cross-level fusion(two-dir)} in \cref{tab:partnet-ablation}.

\myparagraph{Bandwidth of mean-shift} We experienced different bandwidth values for the mean-shift algorithm: $0.05,0.10,0.20$, with \emph{cross-level fusion$^\dagger$} setting. Their performance results are slightly different, as shown in the first three rows of ~\cref{tab:mean-shift-ablation}. Mean-shift with bandwidth $0.10$ performed better than the other two choices. Therefore, we used $0.10$ by default.

\begin{table}[t]
    \centering
    \resizebox{0.8\columnwidth}{!}{
        \begin{tabular}{cc|*{18}{c}}
            \toprule
            bandwidth   & $\lambda$      & $AP_{25}$    & $AP_{50}$ & $AP_{75}$    & s-$AP_{50}$ \\\midrule
            0.05        & -              & 70.5         & 61.8      & 48.2         & 75.0         \\
            0.10        & -              & \textbf{71.3}         & \textbf{63.1}      & \textbf{48.6}         & \textbf{75.2}         \\
            0.20        & -              & 71.1         & 62.4      & \textbf{48.6}         & 74.4         \\\midrule
            0.10        & 0.025          & 71.9         & 64.0      & \textbf{49.7}         & 76.0         \\
            0.10        & 0.050          & \textbf{72.1}         & \textbf{64.1}      & \textbf{49.7}         & \textbf{76.1}         \\
            0.10        & 0.075          & 70.0         & 61.9      & 47.5         & 74.8         \\
            \bottomrule
        \end{tabular}}
    \caption{Bandwidth and $\lambda$ selection. The first three rows are our results for \emph{cross-level fusion$^\dagger$} with different bandwidths. The last three rows are the results for \emph{cross-level fusion} with different $\lambda$ settings.}
    \label{tab:mean-shift-ablation} \vspace{-4mm}
\end{table}

\myparagraph{Choices of $\lambda$} With the default bandwidth of the mean-shift algorithm, we experienced several choices of $\lambda$ for \cref{eq:shiftcenter}: 0.025, 0.050, 0.075, under \emph{cross-level fusion}. The last three rows of ~\cref{tab:mean-shift-ablation} show the results. $\lambda=0.050$ achieved the best result, while larger $\lambda$ could damage the centerness of the shifted points and did not comply with the predefined bandwidth. According to our empirical study, $\lambda$ was set to 0.050 by default.


\subsection{Instance segmentation on indoor scenes}\label{subsec:indoor}

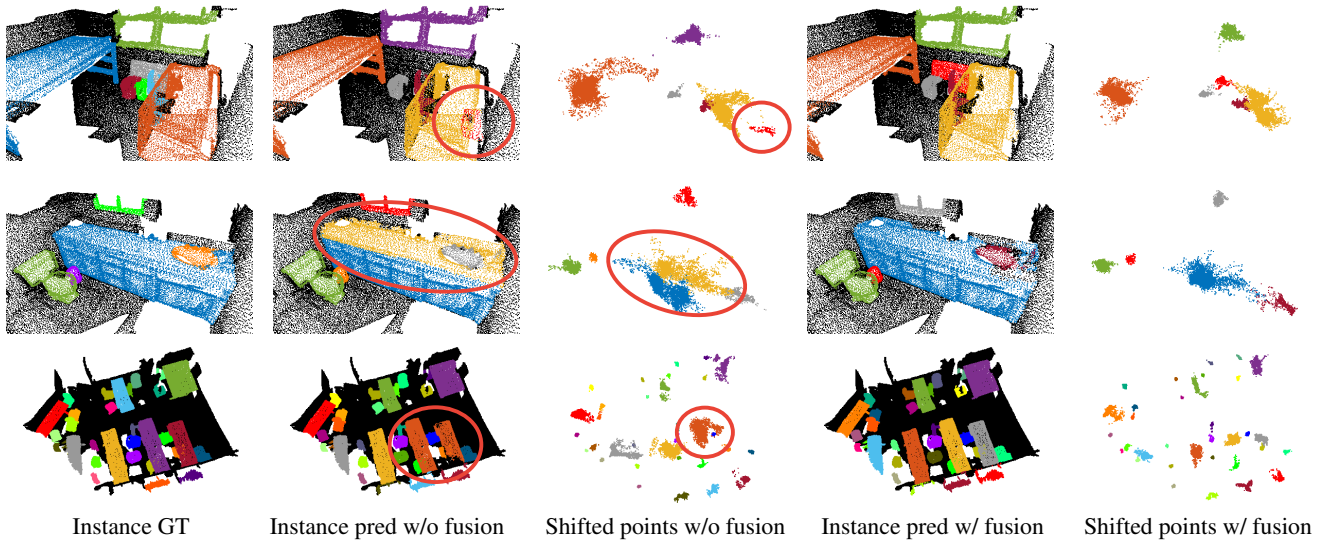
\begin{figure*}[t]
\centering
  \begin{overpic}[width=\linewidth]{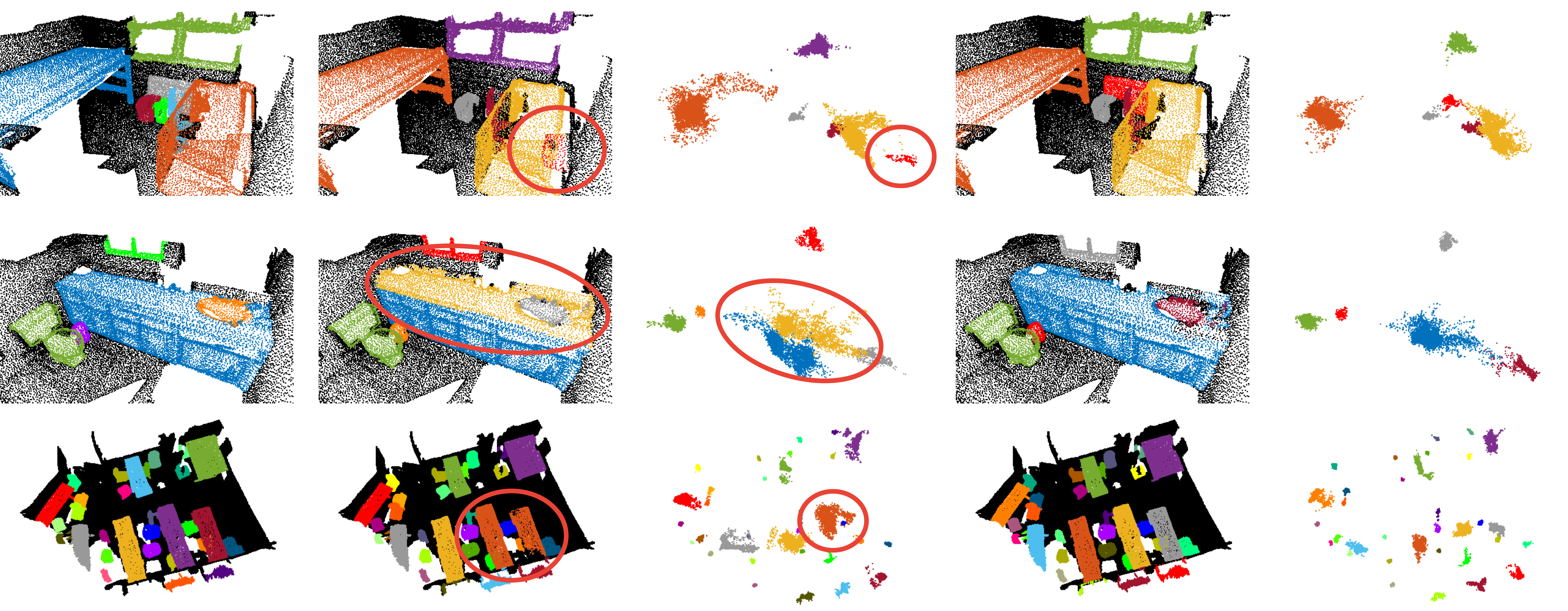}
    \put(5,-2){\small{Instance GT}}
    \put(20,-2){\small{Instance pred w/o fusion}}
    \put(41,-2){\small{Shifted points w/o fusion}}
    \put(62,-2){\small{Instance pred w/ fusion}}
    \put(82,-2){\small{Shifted points w/ fusion}}
   \end{overpic}
  \vspace{1mm}
  \caption{Visual comparison of instance segmentation on the validation set of ScanNet. Without the fusion module, the shifted points are more dispersive and result in wrong instance segmentation results, as shown in the red circles. Our fusion module can help to get more accurate offsets, and the compact shifted points can get better instance clustering.
  }
  \label{fig:scene_comparison} \vspace{-2mm}
\end{figure*}

\myparagraph{Datasets}
The ScanNet~\cite{dai2017scannet} dataset contains 1613 scans with annotations of 3D object instances. 
Instance segmentation was evaluated on 18 object categories. We report the results on the validation set. The S3DIS~\cite{S3DIS} dataset has 272 scenes with 13 categories. It was collected from six large-scale areas, covering more than 6000 $m^2$ with more than 273 million points. We report the performance on both Area-5 and 6-fold sets. \looseness=-1

\myparagraph{Evaluation metrics}
For ScanNet, we use the widely-adopted evaluation metric, $mAP$; $AP_{25}$ and $AP_{50}$ denote AP scores with the IoU threshold of 0.25 and 0.5, respectively. In addition, AP averages the scores with the IoU threshold set from 0.5 to 0.95, with a step size of 0.05. For S3DIS, we use the metrics proposed by ~\cite{ASIS}: mCov, mWCov, mPrec, and mRec. mCov is the mean instance-wise IoU. mWCov is the weighted version of mCov, where the weights are determined by the sizes of each instance. mPrec and mRec denote the mean precision and recall with an IoU threshold of 0.5. In both datasets, we also report the semantic segmentation metric mIoU for reference.

\myparagraph{Experiment setup}
To demonstrate the efficiency of our instance feature fusion module and its applicability to different network designs, we integrated our single-level fusion module into some recent instance segmentation frameworks, which have both the semantic segmentation branch and the instance feature branch: PointGroup~\cite{jiang2020pointgroup}, DyCo3D~\cite{DyCo3D}, HAIS ~\cite{chen2021hierarchical}, ASIS~\cite{ASIS} and JSNet~\cite{zhao2020jsnet}. The settings of the original frameworks, such as loss functions, clustering algorithms, and training protocols, were kept. Our multi- or cross-level fusion is not used here as there are no multi-level instances on the indoor scene datasets. On the ScanNet dataset, we used the original frameworks of PointGroup, DyCo3D, and HAIS as baselines and inserted our fusion module to help in network training. As the work of HAIS and DyCo3D leveraged pretrained network weights to initialize the network weights to obtain high performance, for a fair comparison, we followed their method and used pretrained weights as initialization to train their networks with our fusion module. In Appendix B, we also provided the comparison without using any pretrained weights. On the S3DIS dataset, we retrained ASIS and JSNet with and without their original fusion modules, and trained the networks by replacing their fusion modules with our fusion module for further comparison.

\begin{table}[t]
    \centering
    \resizebox{0.7\columnwidth}{!}
    {
        \begin{tabular}{l|cccc}
            \toprule
            Method                   & $AP$ & $AP_{50}$ & $AP_{25}$  & mIoU\\\midrule
            PointGroup                 & 35.2          & 57.1       & 71.4    & 67.3     \\
            PointGroup$^*$         & \textbf{37.6}          & \textbf{58.7}       & \textbf{71.8}   & \textbf{67.6}       \\\midrule
            DyCo3D                     & 35.5          & 57.6       & 72.9      & \textbf{69.5}    \\
            DyCo3D$^*$             & \textbf{36.6}          & \textbf{58.3}       & \textbf{73.2}       & \textbf{69.5}   \\\midrule
            HAIS                       & 44.1          & 64.4       & 75.7      & 72.3   \\
            HAIS$^*$              & \textbf{44.9}          & \textbf{64.9}       & \textbf{75.9}    & \textbf{72.4}      \\
            \bottomrule
        \end{tabular}}
    \caption{Quantitative comparison on ScanNet~\cite{dai2017scannet} validation set. Our fusion module is added to each network (marked with $*$) and exhibits consistent performance improvements. The results of other methods are from their released models and checkpoints. We used their pre-trained weights for initialization and training of the whole network with our fusion module.}
    \label{tab:scannet_results} 
\end{table}

\begin{table}[t]
    \centering
    \resizebox{\columnwidth}{!}{
    \begin{tabular}{l|ccccc}
    \toprule
     Method  & mCov &  mWCov & mPrec & mRec & mIoU\\
    \midrule
    b-ASIS & 45.4(49.0) & 48.6(53.0) & 53.7(58.8) & 42.9(47.3) & 52.0(58.4)\\
    ASIS & 45.8(49.4) & 48.9(53.3) & 54.7(59.5) & \textbf{43.6}(47.4) & 52.3(58.8)\\
    b-ASIS$^*$ & \textbf{46.1}(\textbf{50.4}) & \textbf{49.2}(\textbf{54.4}) & \textbf{55.4}(\textbf{63.0}) & 43.4(\textbf{50.2}) & \textbf{53.1}(\textbf{59.3})\\\midrule
    b-JSNet & 47.9(50.8) & 50.7(54.8) & 55.6(60.7) & 44.8(49.7) & 53.5(59.5)\\
    JSNet & 48.8(51.7) & 51.6(55.5) & 56.6(61.1) & 46.1(50.6) & 53.9(59.9)\\
    b-JSNet$^*$ & \textbf{49.5}(\textbf{51.9}) & \textbf{52.6}(\textbf{55.8}) & \textbf{58.6}(\textbf{63.1}) & \textbf{46.6}(\textbf{51.0}) & \textbf{54.7}(\textbf{60.4})\\
    \bottomrule
    \end{tabular}
    }
    \caption{Quantitative comparison on S3DIS~\cite{S3DIS}. b-ASIS is the baseline of ASIS, \ie without the ASIS feature fusion module. Similarly, b-JSNet is the baseline of JSNet. We added our fusion module to each method marked with $*$. The number before parentheses is the metric on Area 5, while the number inside parentheses is the metric on 6-fold cross-validation.
    }
    \label{tab:S3DIS_results}
    \vspace{-4mm}
\end{table}

\begin{table}[t]
    \centering
    \resizebox{0.6\columnwidth}{!}{
        \begin{tabular}{l|c}
            \toprule
            Method                  & Inference time (msec) \\\midrule
            PointGroup                 & \makebox[7ex][l]{428}             \\
            PointGroup$^*$         & \makebox[7ex][l]{439(+11)}         \\\midrule
            DyCo3D                     & \makebox[7ex][l]{392}       \\
            DyCo3D$^*$             & \makebox[7ex][l]{400(+8)}    \\\midrule
            HAIS                       & \makebox[7ex][l]{375}            \\
            HAIS$^*$              & \makebox[7ex][l]{388(+13)}      \\
            \midrule\midrule
            b-ASIS  & \makebox[8ex][l]{3405} \\
            ASIS &  \makebox[8ex][l]{5058(+1653)} \\
            b-ASIS$^*$ & \makebox[8ex][l]{3646(+241)} \\ \midrule
            b-JSNet  & \makebox[8ex][l]{4138} \\
            JSNet &  \makebox[8ex][l]{4256(+118)} \\
            b-JSNet$^*$ & \makebox[8ex][l]{4192(+54)} \\
            \bottomrule
        \end{tabular}}
    \caption{Average inference time for a 3D scan. Methods using our fusion module are marked with $*$. The first three methods are measured on ScanNet validation set and the last two methods are measured on Area 5 of S3DIS. The runtime was measured on Tesla V100 GPU.}
    \label{tab:runtime_analysis} \vspace{-2mm}
\end{table}

\myparagraph{Performance report and time analysis}
\cref{tab:scannet_results} shows the performance results of PointGroup, DyCo3D, and HAIS with and without our fusion module on the validation set of ScanNet. Our fusion module consistently improved these methods: $+2.4$, $+1.1$ and $+0.8$ points on $AP$,  and $+1.6$, $+0.7$ and $+0.5$ points on $AP_{50}$. In \cref{fig:scene_comparison}, we present some instance segmentation results by HAIS with and without our fusion module. Without our fusion module, the shifted points have a larger distribution which can lead to wrong clustering results, as highlighted by the red circles. With our fusion module, the shifted points are closer to their instance centers, which helps to achieve more accurate clustering results.

On S3DIS, we retrained ASIS and JSNet with and without their original fusion modules, and we also integrated our fusion module with their base networks. As reported in  \cref{tab:S3DIS_results}, the improvement of our fusion module outperformed their original fusion modules.

On the above experiments, the additional inference time caused by our fusion module for each method was small compared to the total time, as reported in \cref{tab:runtime_analysis}. The additional time of our fusion is also smaller than the fusion modules of ASIS and JSNet. We conclude that our fusion module is a lightweight and an effective plugin to improve the performance of other methods.

\section{Conclusion} \label{sec:conclusion} 
We present a novel semantic segmentation-assisted instance feature fusion scheme and an improved instance clustering method via the semantic region center for multi-level 3D part instance segmentation. Our method explicitly utilizes the inherent relationship between semantic segmentation and part instances considering their hierarchy. Its efficacy is well demonstrated on a challenging 3D shape dataset --- PartNet. Our feature fusion scheme also integrates well with other state-of-the-art 3D indoor-scene instance segmentation models, which it  consistently improve on ScanNet and S3DIS.

\myparagraph{Limitation} In our algorithm for PartNet, the bandwidth of the mean-shift algorithm and the shift parameter $\lambda$ were set empirically. Devising a differentiable clustering algorithm with trainable bandwidth and $\lambda$ for end-to-end training would help improve the instance segmentation accuracy further. The approach of taking mean-shift iterations as differentiable recurrent functions ~\cite{sharma2020parsenet} is a promising direction.

\section*{Appendices}
\begin{figure*}[t]
\centering
  \begin{overpic}[width=0.8\linewidth]{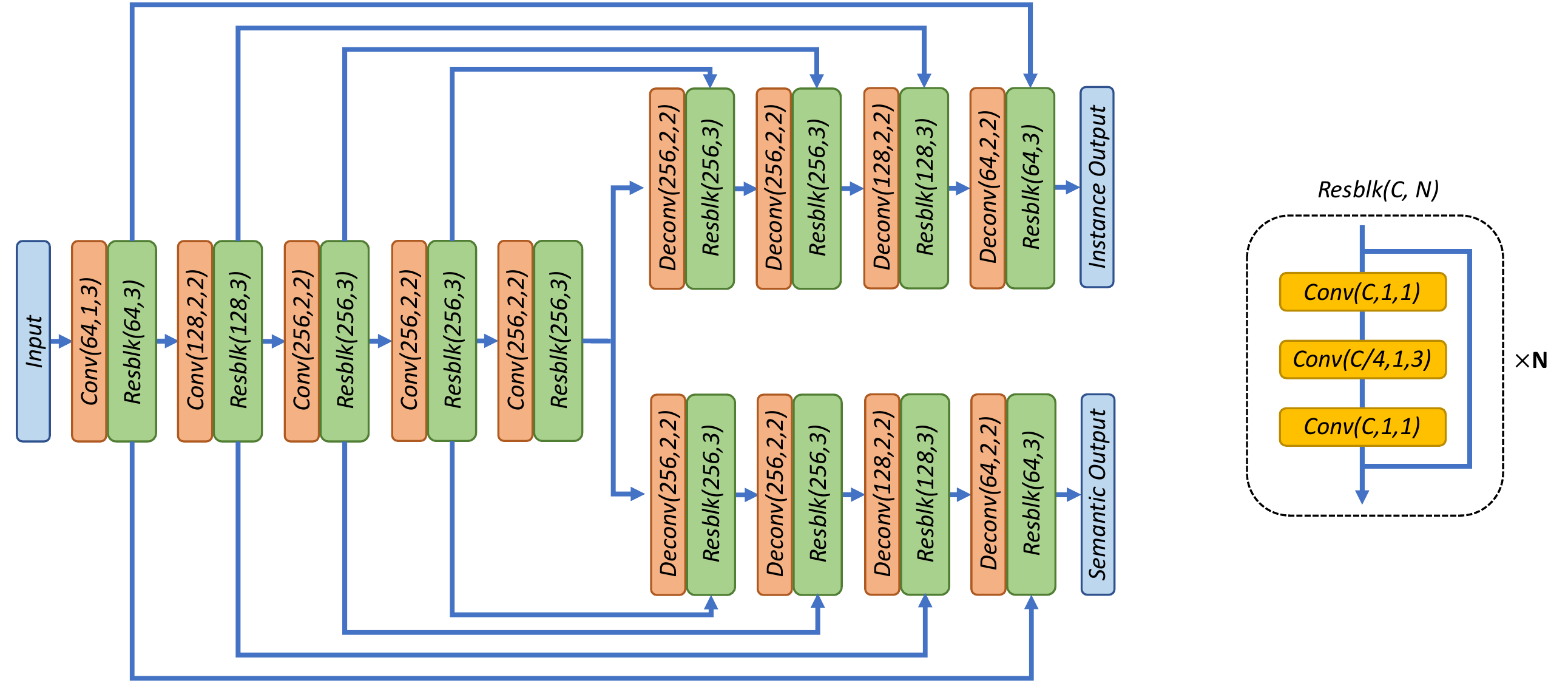}
   \end{overpic}
  \caption{
    O-CNN-based U-Net structure for instance segmentation on the PartNet dataset. $Conv(C, S, K)$ and $Deconv(C, S, K)$ represent octree-based convolution and deconvolution. $C, S, K$ are the output channel number, stride, and kernel size.
  }
  \label{fig:unet} \vspace{-5mm}
\end{figure*}

\subsection*{A. U-Net Structure} \label{appendix:unet}

We used an O-CNN-based U-Net structure with two decoders as our base network. The encoder and decoders have five levels of domain resolution, and the maximum depth of the octree is \num{6}, as illustrated in \cref{fig:unet}.

\subsection*{B. Training from Scratch in ScanNet}
\label{appendix:scratch}
For the methods of PointGroup~\cite{jiang2020pointgroup}, DyCo3D~\cite{DyCo3D} and HAIS~\cite{chen2021hierarchical}, we trained their networks using the default setting of their released codes from scratch with and without our fusion module.
The results in \cref{tab:scannet_results_scratch} show that our fusion module led to consistent improvements. Note that all methods trained from scratch are inferior to their versions using pretrained weights.

\begin{table}[h]
    \centering
    \resizebox{0.8\columnwidth}{!}{
        \begin{tabular}{l|cccc}
            \toprule
            Method                    & $AP$ & $AP_{50}$ & $AP_{25}$  & mIoU\\\midrule
            PointGroup                 & 33.6          & 55.4       & 70.0    & 67.1     \\
            PointGroup$^*$         & \textbf{34.4}          & \textbf{56.1}       & \textbf{71.7}   & \textbf{67.3}       \\\midrule
            DyCo3D                     & 32.5          & 53.0       & 69.0      & 67.2    \\
            DyCo3D$^*$             & \textbf{34.5}          & \textbf{55.8}       & \textbf{70.7}       & \textbf{67.6}   \\\midrule
            HAIS                       & 42.5          & 61.7       & 73.5      & 71.0   \\
            HAIS$^*$              & \textbf{43.1}          & \textbf{62.8}       & \textbf{74.5}    & \textbf{71.4}      \\
            \bottomrule
        \end{tabular}}
    \caption{Quantitative comparison on ScanNet~\cite{dai2017scannet} validation set. Our fusion module is added to each network (marked with $*$) and exhibits consistent performance improvements. The other methods are trained from scratch using their released codes. The networks with our fusion module are also trained from scratch.}
    \label{tab:scannet_results_scratch} \vspace{-2mm}
\end{table}

\noindent\textbf{Remark}: The above networks trained from scratch do not reproduce the performance of the released checkpoints of these works. The authors of DyCo3D and HAIS responded that their released checkpoints used other pre-trained network weights and were not trained from scratch. 

\subsection*{C. Evaluation and Visualization in PartNet} \label{appendix:partnet}

\begin{figure*}[t]
\centering
  \begin{overpic}[width=0.8\linewidth]{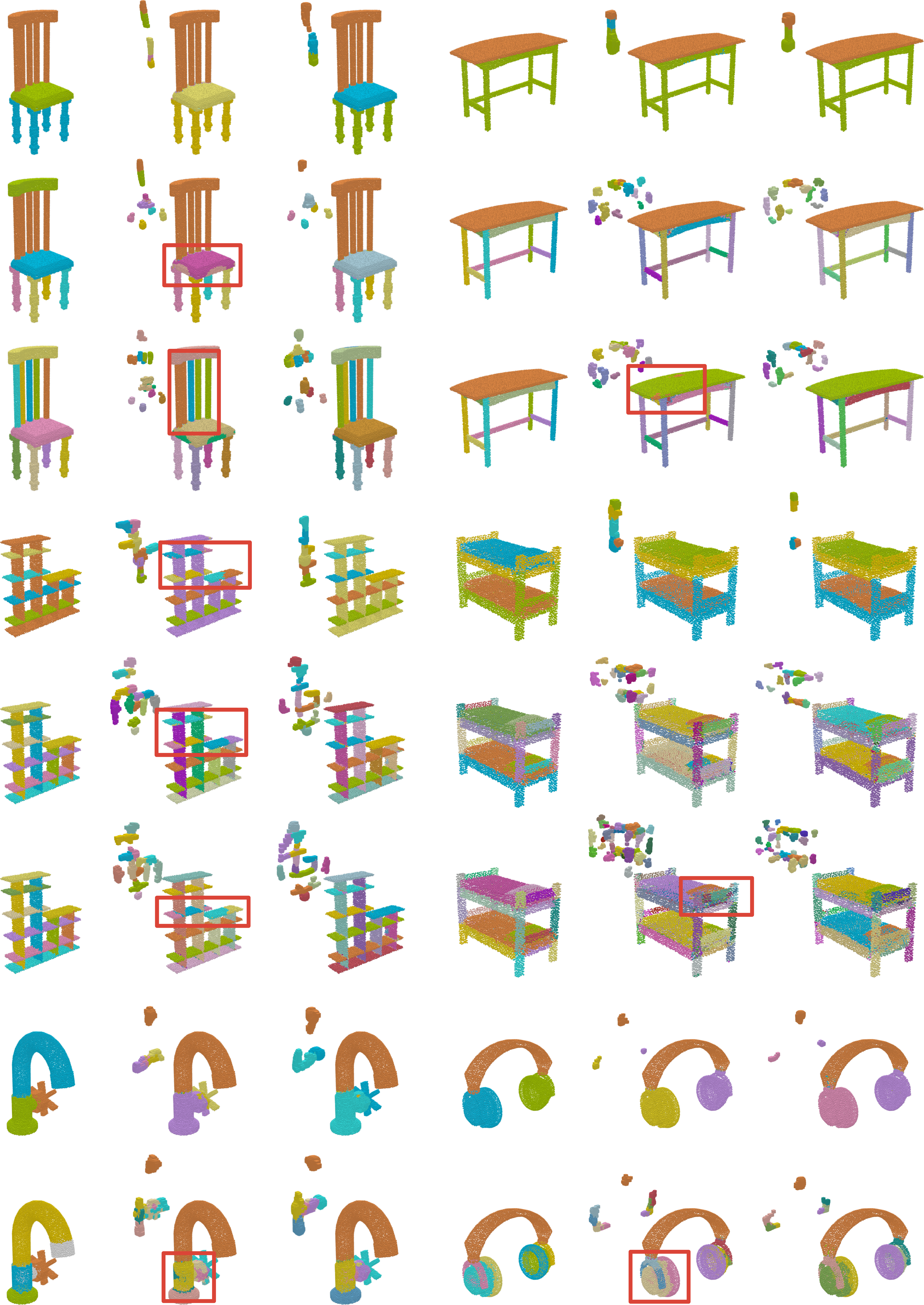}
    \put(2,-2){\small(a)}
    \put(14.5,-2){\small(b)}
    \put(27,-2){\small(c)}
    \put(38.5,-2){\small(a)}
    \put(52,-2){\small(b)}
    \put(65.75,-2){\small(c)}
    \put(-6,94){\small Coarse}
    \put(-6,81){\small Middle}
    \put(-5,68){\small Fine}
    \put(-6,55){\small Coarse}
    \put(-6,42){\small Middle}
    \put(-5,29){\small Fine}
    \put(-6,17){\small Coarse}
    \put(-5,4){\small Fine}
   \end{overpic}
   \vspace{2mm}
  \caption{Visual comparison of part instance segmentation on the test set PartNet. Part instances at each level are colored with random colors. (a) Ground truth instance parts. (b) Results of our \textit{multi-level baseline$^{\dagger}$}. (c) Results of our \textit{cross-level fusion}. The corresponding shifted points are rendered on the top-left of each instance segmentation image. Red boxes represent wrong instance results.
  }
  \label{fig:comparison-3level} \vspace{-2mm}
\end{figure*}
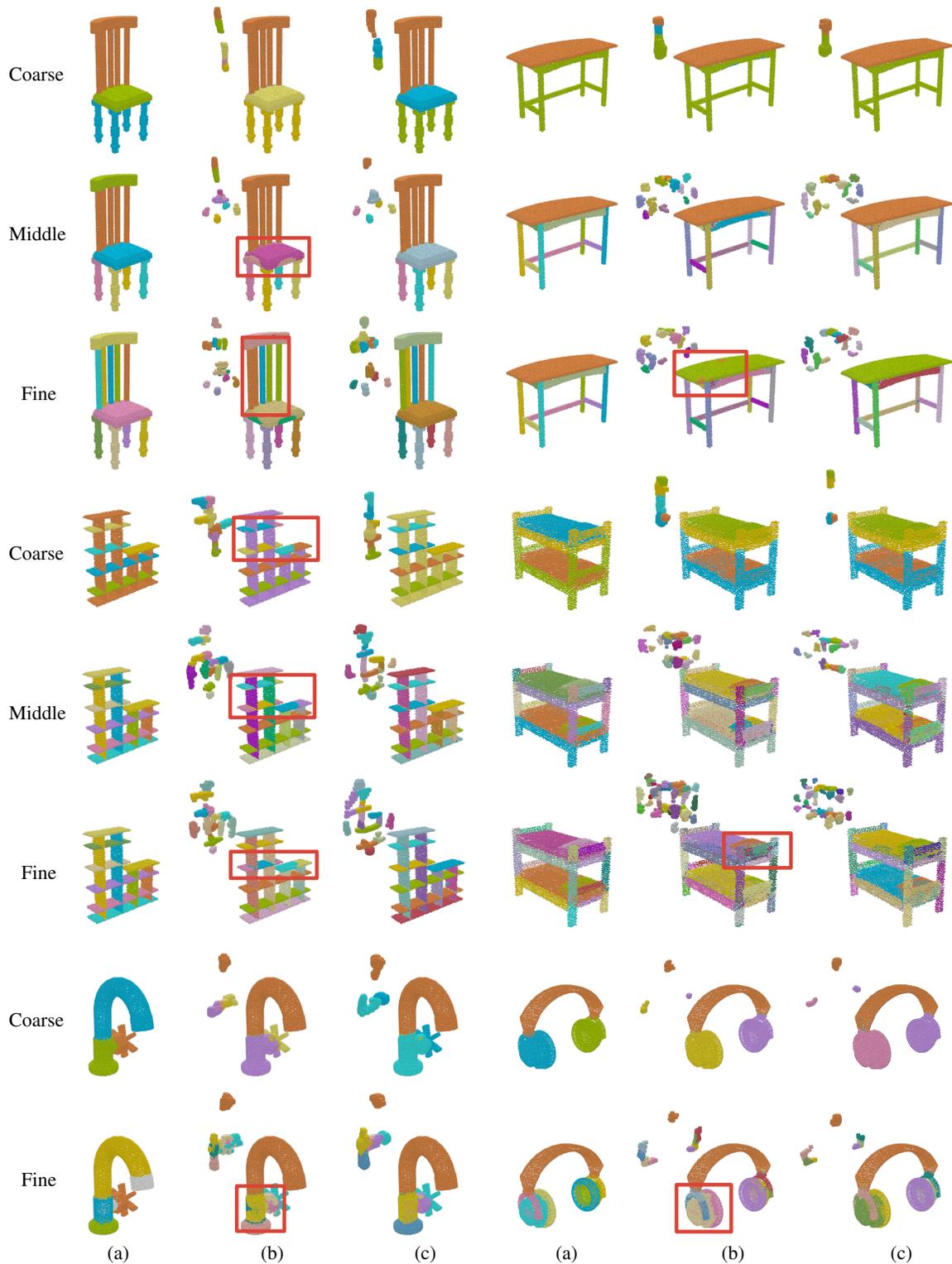

We report $AP_{25}$, $AP_{75}$ and s-$AP_{50}$ on the 24 shape categories of PartNet in \cref{tab:partnet-othermetric}.
In \cref{fig:comparison-3level}, we illustrate the \emph{multi-level baseline} and \emph{cross-level fusion} instance segmentation results. Our fusion module helps obtain more compact and distinguishable instance centers and yielded better instance segmentation results.

\begin{table*}[t]

    \resizebox{\textwidth}{!}{%

        \begin{tabular}{@{}c|c|c|*{24}{c}@{}}
            \toprule
             &  Level   & Avg           & \rotatebox[origin=lB]{60}{Bag} & \rotatebox[origin=lB]{60}{Bed} & \rotatebox[origin=lB]{60}{Bottle} & \rotatebox[origin=lB]{60}{Bowl} & \rotatebox[origin=lB]{60}{Chair} & \rotatebox[origin=lB]{60}{Clock} & \rotatebox[origin=lB]{60}{Dish} & \rotatebox[origin=lB]{60}{Disp} & \rotatebox[origin=lB]{60}{Door} & \rotatebox[origin=lB]{60}{Ear} & \rotatebox[origin=lB]{60}{Faucet} & \rotatebox[origin=lB]{60}{Hat} & \rotatebox[origin=lB]{60}{Key} & \rotatebox[origin=lB]{60}{Knife} & \rotatebox[origin=lB]{60}{Lamp} & \rotatebox[origin=lB]{60}{Laptop} & \rotatebox[origin=lB]{60}{Micro} & \rotatebox[origin=lB]{60}{Mug} & \rotatebox[origin=lB]{60}{Fridge} & \rotatebox[origin=lB]{60}{Scis} & \rotatebox[origin=lB]{60}{Stora} & \rotatebox[origin=lB]{60}{Table} & \rotatebox[origin=lB]{60}{Trash} & \rotatebox[origin=lB]{60}{Vase} \\ \toprule

            \multirow{4}{*}{\rotatebox[origin=c]{90}{PartNet~\cite{Mo2019}}}
             & Coarse   & 70.2 &  \textbf{89.4} &  \textbf{82.3} & 65.2 & 63.1 & 78.1 & 48.0 & 79.1 & 97.1 & 64.9 & 64.6 & 77.3 & 73.9 & 58.9 & 59.2 & 42.5 &  \textbf{100.0} & 50.0 & 92.9 & 50.0 & 96.3 & 57.7 &  59.3 & 82.7 &  52.6                            \\
             & Middle   & 46.7 & - & 44.5 & - & - & 43.0 & - & 71.3 & - &  49.3 & - & - & - & - & - & 32.2 & - & 51.2 & - & 45.2 & - & 46.7 & 36.5 & - & -                               \\
             & Fine   & 45.6 & - & 29.0 & 52.6 & - & 35.3 & 39.6 & 59.9 & 89.3 & 27.1 & 56.9 & 55.0 & - & - & 49.0 & 22.6 & - & 56.9 & - & 35.6 & - & 36.3 & 28.6 & 44.8 &  57.0                            \\ \cmidrule{2-27}
             & Avg & 62.8 &  89.4 & 51.9 & 58.9 & 63.1 & 52.1 & 43.8 & 70.1 & 93.2 & 47.1 & 60.8 & 66.2 & 73.9 & 58.9 & 54.1 & 32.4 &  \textbf{100.0} & 52.7 & 92.9 & 43.6 & 96.3 & 46.9 &  41.5 & 63.8 &  54.8                            \\ \toprule
            \multirow{4}{*}{\rotatebox[origin=c]{90}{PE~\cite{zhang2021}}}
             & Coarse   &  72.7 & 82.8 & 79.6 &  65.6 &  72.0 &  82.8 &  49.1 &  83.8 &  98.3 &  75.5 &  74.3 &  83.2 &  79.5 &  \textbf{59.9} &  \textbf{78.8} &  \textbf{45.2} &  \textbf{100.0} &  50.5 &  95.4 &  51.6 &  \textbf{96.9} &  60.9 & 44.6 &  82.9 & 51.1                            \\
             & Middle   &  51.4 & - &  55.4 & - & - &  47.1 & - &  78.0 & - & 48.1 & - & - & - & - & - &  \textbf{39.3} & - &  54.4 & - &  48.8 & - &  53.7 &  37.7 & - & -                               \\
             & Fine   &  51.6 & - &  44.4 &  57.2 & - &  43.2 &  45.7 &  64.8 &  90.7 &  34.6 &  59.3 &  67.2 & - & - &  53.0 &  26.0 & - &  60.0 & - &  \textbf{51.5} & - &  44.4 &  31.7 &  50.0 & 53.9                            \\ \cmidrule{2-27}
             & Avg &  66.5 & 82.8 &  59.8 &  61.4 &  72.0 &  57.7 &  47.4 &  75.6 &  94.5 &  52.7 &  66.8 &  75.2 &  79.5 &  \textbf{59.9} &  \textbf{65.9} &  36.8 &  \textbf{100.0} &  55.0 &  95.4 &  50.6 &  \textbf{96.9} &  53.0 & 38.0 &  66.5 & 52.5                            \\ \toprule

            \multirow{4}{*}{\rotatebox[origin=c]{90}{ Ours}}
             & Coarse & \textbf{78.1}	& 83.8 & 72.7 &	\textbf{68.2} &	\textbf{84.2} &	\textbf{87.3} &	\textbf{58.5} &	\textbf{87.2} &	\textbf{99.0} &	\textbf{82.8} &	\textbf{80.5} &	\textbf{88.2} &	\textbf{87.9} &	53.2 &	71.3 &	43.9 &	\textbf{100.0} &	\textbf{83.4} &	\textbf{96.7} &	\textbf{61.3} &	96.0 &	\textbf{70.3} &	\textbf{77.1} &	\textbf{85.8} &	\textbf{54.3 }             \\
             & Middle   & \textbf{62.1} & - &	     	\textbf{67.3} & - & - &	     	     	\textbf{54.0} & - &	     	\textbf{82.6} & - &	     	\textbf{67.1} & - & - & - & - & - &	     	     	     	     	     	38.8 & - &	     	\textbf{80.9} & - &	     	\textbf{60.5} & - &	     	\textbf{64.4} &	\textbf{43.4} & - & -	     	     
                               \\
             & Fine   & \textbf{58.4} & - &	     	\textbf{58.8} &	\textbf{62.9} & - &	     	\textbf{47.0} &	\textbf{49.9} &	\textbf{70.1} &	\textbf{93.4} &	\textbf{52.0} &	\textbf{65.4} &	\textbf{70.8} & - & - &	     	     	\textbf{54.3} &	\textbf{30.0} & - &	     	\textbf{72.2} & - &	     	51.2 & - &	     	\textbf{53.8} &	\textbf{36.7} &	\textbf{62.0} &	\textbf{61.8}
                            \\ \cmidrule{2-27}
             & Avg & \textbf{72.1} &	83.8 &	\textbf{66.3} &	\textbf{65.6} &	\textbf{84.2} &	\textbf{62.8} &	\textbf{54.2} &	\textbf{80.0} &	\textbf{96.2} &	\textbf{67.3} &	\textbf{73.0} &	\textbf{79.5} &	\textbf{87.9} &	53.2 &	62.8 &	\textbf{37.6} &	\textbf{100.0} &	\textbf{78.8} &	\textbf{96.7} &	\textbf{57.7} &	96.0 &	\textbf{62.8} &	\textbf{52.4} &	\textbf{73.9} &	\textbf{58.1}
                   \\ 
            \bottomrule   \multicolumn{27}{c}{} \\
            \multicolumn{27}{c}{\LARGE \textbf{$AP_{25}$ }} \\
            \multicolumn{27}{c}{} \\ \multicolumn{27}{c}{} \\
                        \toprule
             &  Level   & Avg           & \rotatebox[origin=lB]{60}{Bag} & \rotatebox[origin=lB]{60}{Bed} & \rotatebox[origin=lB]{60}{Bottle} & \rotatebox[origin=lB]{60}{Bowl} & \rotatebox[origin=lB]{60}{Chair} & \rotatebox[origin=lB]{60}{Clock} & \rotatebox[origin=lB]{60}{Dish} & \rotatebox[origin=lB]{60}{Disp} & \rotatebox[origin=lB]{60}{Door} & \rotatebox[origin=lB]{60}{Ear} & \rotatebox[origin=lB]{60}{Faucet} & \rotatebox[origin=lB]{60}{Hat} & \rotatebox[origin=lB]{60}{Key} & \rotatebox[origin=lB]{60}{Knife} & \rotatebox[origin=lB]{60}{Lamp} & \rotatebox[origin=lB]{60}{Laptop} & \rotatebox[origin=lB]{60}{Micro} & \rotatebox[origin=lB]{60}{Mug} & \rotatebox[origin=lB]{60}{Fridge} & \rotatebox[origin=lB]{60}{Scis} & \rotatebox[origin=lB]{60}{Stora} & \rotatebox[origin=lB]{60}{Table} & \rotatebox[origin=lB]{60}{Trash} & \rotatebox[origin=lB]{60}{Vase} \\ \toprule

            \multirow{4}{*}{\rotatebox[origin=c]{90}{PartNet~\cite{Mo2019}}}
             & Coarse   & 47.4 & 39.7 & 14.6 & \textbf{60.6} & 41.4 & 58.3 & 28.8 & 58.3 & 84.7 & 35.6 & 49.1 & 48.2 & 66.3 & 10.7 & 48.7 & \textbf{29.6} & \textbf{98.0} & 47.8 & 76.1 & 50.0 & 35.1 & 29.9 & 43.2 & 42.2 & 40.5                            \\
             & Middle   & 22.0 & - & 4.2 & - & - & 21.4 & - & 37.2 & - & 22.4 & - & - & - & - & - & 19.6 & - & 32.1 & - & 16.7 & - & 22.8 & 22.0 & - & -                               \\
             & Fine   & 23.5 & - & 3.9 & 37.9 & - & 16.6 & 17.6 & 29.8 & 63.2 & 8.1 & 27.6 & 25.8 & - & - & 31.0 & 13.6 & - & 23.9 & - & 12.1 & - & 18.2 & 16.4 & 19.7 & 34.5                            \\ \cmidrule{2-27}
             & Avg & 38.9 & 39.7 & 7.6 & 49.2 & 41.4 & 32.1 & 23.2 & 41.7 & 73.9 & 22.0 & 38.4 & 37.0 & 66.3 & 10.7 & 39.8 & 20.9 & \textbf{98.0} & 34.6 & 76.1 & 26.3 & 35.1 & 23.6 & \textbf{27.2} & 31.0 & 37.5                            \\ \toprule
            \multirow{4}{*}{\rotatebox[origin=c]{90}{PE~\cite{zhang2021}}}
             & Coarse   &  50.0 & 40.3 & 13.3 & 60.2 & 60.2 & 59.3 & 28.2 & 61.9 & 90.6 & 39.1 & \textbf{59.6} & 54.2 & 69.3 & 7.4 & \textbf{65.7} & 28.5 & \textbf{98.0} & 47.9 & 77.1 & 50.5 & 42.8 & 30.1 & 34.8 & 40.7 & 41.1                            \\
             & Middle   &  23.8 & - & 7.1 & - & - & 22.8 & - & 37.4 & - & 21.3 & - & - & - & - & - & 22.0 & - & \textbf{35.5} & - & 20.6 & - & 26.1 & 21.4 & - & -                               \\
             & Fine   &  25.7 & - & 7.3 & \textbf{38.8} & - & 20.5 & 17.2 & 30.0 & 66.8 & 10.8 & 28.2 & 33.2 & - & - & 31.5 & 14.1 & - & 25.6 & - & 17.1 & - & 21.0 & 17.4 & 19.4 & 38.0                            \\ \cmidrule{2-27}
             & Avg &  41.7 & 40.3 & 9.2 & \textbf{49.5} & 60.2 & 34.2 & 22.7 & 43.1 & 78.7 & 23.7 & \textbf{43.9} & 43.7 & 69.3 & 7.4 & 48.6 & 21.5 & \textbf{98.0} & \textbf{36.4} & 77.1 & 29.4 & 42.8 & 25.7 & 24.5 & 30.0 & 39.6                            \\ \toprule

            \multirow{4}{*}{\rotatebox[origin=c]{90}{ Ours}}
             & Coarse & \textbf{57.7} &  \textbf{61.1} &  \textbf{22.0} &  44.8 &  \textbf{66.1} &  \textbf{70.6} &  \textbf{37.4} &  \textbf{65.0} &  \textbf{91.7} &  \textbf{52.3} &  55.5 &  \textbf{65.5} &  \textbf{72.4} &  \textbf{44.4} &  62.1 &  28.6 &  \textbf{98.0} &  \textbf{49.0} &  \textbf{87.9} &  \textbf{54.0} &  \textbf{62.8} &  \textbf{37.7} &  \textbf{48.7} &  \textbf{61.0} &  \textbf{45.3} \\
             & Middle   & \textbf{31.2} & - &        \textbf{19.1} & - & - &              \textbf{32.4} & - &        \textbf{47.7} & - &        \textbf{32.5} & - & - & - & - & - &                                \textbf{23.6} & - &        33.5 & - &        \textbf{29.4} & - &        \textbf{36.6} &  \textbf{25.9} & - & -    \\
             & Fine   & \textbf{31.6} & - &        \textbf{16.9} &  33.2 & - &        \textbf{25.4} &  \textbf{19.6} &  \textbf{36.3} &  \textbf{74.5} &  \textbf{21.9} &  \textbf{30.8} &  \textbf{44.8} & - & - &              \textbf{35.5} &  \textbf{16.4} & - &        \textbf{25.9} & - &        \textbf{27.8} & - &        \textbf{30.4} &  \textbf{20.4} &  \textbf{33.9} &  \textbf{43.0}  \\ \cmidrule{2-27}
             & Avg & \textbf{49.7} & \textbf{61.1} &  \textbf{19.3} &  39.0 &  \textbf{66.1} &  \textbf{42.8} &  \textbf{28.5} &  \textbf{49.7} &  \textbf{83.1} &  \textbf{35.6} &  43.2 &  \textbf{55.2} &  \textbf{72.4} &  \textbf{44.4} &  \textbf{48.8} &  \textbf{22.9} &  \textbf{98.0} &  36.1 &  \textbf{87.9} &  \textbf{37.1} &  \textbf{62.8} &  \textbf{34.9} &  \textbf{31.7} &  \textbf{47.5} &  \textbf{44.2} \\
            \bottomrule   \multicolumn{27}{c}{} \\
            \multicolumn{27}{c}{\LARGE \textbf{$AP_{75}$ }} \\
            
                        \multicolumn{27}{c}{} \\ \multicolumn{27}{c}{} \\

                        \toprule
             &  Level   & Avg           & \rotatebox[origin=lB]{60}{Bag} & \rotatebox[origin=lB]{60}{Bed} & \rotatebox[origin=lB]{60}{Bottle} & \rotatebox[origin=lB]{60}{Bowl} & \rotatebox[origin=lB]{60}{Chair} & \rotatebox[origin=lB]{60}{Clock} & \rotatebox[origin=lB]{60}{Dish} & \rotatebox[origin=lB]{60}{Disp} & \rotatebox[origin=lB]{60}{Door} & \rotatebox[origin=lB]{60}{Ear} & \rotatebox[origin=lB]{60}{Faucet} & \rotatebox[origin=lB]{60}{Hat} & \rotatebox[origin=lB]{60}{Key} & \rotatebox[origin=lB]{60}{Knife} & \rotatebox[origin=lB]{60}{Lamp} & \rotatebox[origin=lB]{60}{Laptop} & \rotatebox[origin=lB]{60}{Micro} & \rotatebox[origin=lB]{60}{Mug} & \rotatebox[origin=lB]{60}{Fridge} & \rotatebox[origin=lB]{60}{Scis} & \rotatebox[origin=lB]{60}{Stora} & \rotatebox[origin=lB]{60}{Table} & \rotatebox[origin=lB]{60}{Trash} & \rotatebox[origin=lB]{60}{Vase} \\ \toprule

            \multirow{4}{*}{\rotatebox[origin=c]{90}{SGPN~\cite{SGPN}}}
             & Coarse   & 72.5  & 62.8  & 38.7  & 76.7  & 83.2  & 91.5  & 41.5  & 81.4  & 91.3  & 71.2  & 81.4  & 82.2  & 71.9  & 23.2  & \textbf{78.0}  & 60.3  & \textbf{100.0}  & 76.2  & 94.3  & 60.6  & 74.9  & 55.0  & 80.1  & 76.1  & 87.1                            \\
             & Middle   & 50.2  & -  & 22.7  & -  & -  & 51.1  & -  & 78.7  & -  & 43.3  & -  & -  & -  & -  & -  & 49.1  & -  & 68.6  & -  & 42.9  & -  & 51.9  & 43.7  & -  & -                               \\
             & Fine   & 50.2  & -  & 17.5  & 66.5  & -  & 42.3  & 40.7  & 59.3  & 83.9  & 29.0  & 60.2  & 61.6  & -  & -  & 55.0  & 37.6  & -  & 53.7  & -  & 30.6  & -  & 45.1  & 37.8  & 50.0  & 82.0                            \\ \cmidrule{2-27}
             & Avg & 64.2  & 62.8  & 26.3  & 71.6  & 83.2  & 61.6  & 41.1  & 73.1  & 87.6  & 47.8  & 70.8  & 71.9  & 71.9  & 23.2  & 66.5  & 49.0  & \textbf{100.0}  & 66.2  & 94.3  & 44.7  & 74.9  & 50.7  & 53.8  & 63.0  & 84.6                            \\ \toprule
            \multirow{4}{*}{\rotatebox[origin=c]{90}{PartNet~\cite{Mo2019}}}
             & Coarse   &   80.3  & 78.4  & 62.2  & 80.8  & 83.8  & \textbf{94.9}  & 74.6  & 81.4  & 94.3  & 76.1  & \textbf{87.1}  & \textbf{86.5}  & 77.8  & 44.5  & 76.6  & 65.0  & \textbf{100.0}  & 79.5  & 95.3  & \textbf{79.0}  & 87.6  & 62.7  & 88.1  & 82.3  & \textbf{89.0}                            \\
             & Middle   &  60.5  & -  & 29.4  & -  & -  & 64.7  & -  & 75.4  & -  & 61.1  & -  & -  & -  & -  & -  & \textbf{56.8}  & -  & 78.2  & -  & 61.7  & -  & 57.4  & \textbf{59.4}  & -  & -                               \\
             & Fine   &  57.7  & -  & 22.1  & 68.3  & -  & 58.4  & 53.7  & 67.5  & 84.8  & 38.0  & 62.4  & 66.8  & -  & -  & \textbf{63.5}  & 45.8  & -  & 54.0  & -  & 45.0  & -  & 52.6  & \textbf{52.5}  & 58.7  & 86.4                            \\ \cmidrule{2-27}
             & Avg &  72.2  & 78.4  & 37.9  & 74.6  & 83.8  & 72.7  & 64.2  & 74.8  & 89.5  & 58.4  & 74.8  & 76.6  & 77.8  & 44.5  & \textbf{70.1}  & 55.8  & \textbf{100.0}  & 70.6  & 95.3  & 61.9  & 87.6  & 57.6  & 66.7  & 70.5  & 87.7                            \\ \toprule

            \multirow{4}{*}{\rotatebox[origin=c]{90}{ Ours}}
             & Coarse & \textbf{83.3} &  \textbf{84.6} &  \textbf{75.8} &  \textbf{91.0} &  \textbf{88.7} &  94.8 &  \textbf{74.9} &  \textbf{86.3} &  \textbf{97.4} &  \textbf{83.4} &  86.3 &  85.7 &  \textbf{80.1} &  \textbf{47.4} &  76.5 &  \textbf{65.8} &  \textbf{100.0} & \textbf{84.7} &  \textbf{96.2} &  75.9 &  \textbf{88.6} &  \textbf{73.0} &  \textbf{90.5} &  \textbf{83.4} &  88.5
 \\
             & Middle   & \textbf{66.2} & - &        \textbf{50.7} & - & - &              \textbf{65.6} & - &        \textbf{81.5} & - &        \textbf{66.5} & - & - & - & - & - &                                54.8 & - &        \textbf{80.9} & - &        \textbf{70.9} & - &        \textbf{66.7} &  58.5 & - & -             
    \\
             & Fine   & \textbf{63.9} & - &        \textbf{39.1} &  \textbf{70.1} & - &        \textbf{59.5} &  \textbf{54.8} &  \textbf{69.5} &  \textbf{89.1} &  \textbf{56.5} &  \textbf{69.5} &  \textbf{73.7} & - & - &              55.6 &  \textbf{47.4} & - &        \textbf{67.2} & - &        \textbf{63.3} & - &        \textbf{63.9} &  51.9 &  \textbf{66.2} &  \textbf{88.4}
  \\ \cmidrule{2-27}
             & Avg & \textbf{76.1} & \textbf{84.6} &  \textbf{55.2} &  \textbf{80.6} &  \textbf{88.7} &  \textbf{73.3} &  \textbf{64.9} &  \textbf{79.1} &  \textbf{93.3} &  \textbf{68.8} &  \textbf{77.9} &  \textbf{79.7} &  \textbf{80.1} &  \textbf{47.4} &  66.1 &  \textbf{56.0} &  \textbf{100.0} & \textbf{77.6} &  \textbf{96.2} &  \textbf{70.0} &  \textbf{88.6} &  \textbf{67.9} &  \textbf{67.0} &  \textbf{74.8} &  \textbf{88.5}
 \\

            \bottomrule   \multicolumn{27}{c}{} \\
            \multicolumn{27}{c}{\LARGE \textbf{s-$AP_{50}$}} \\

        \end{tabular}
    }
    \caption{Part instance segmentation results on the test set of PartNet~\cite{Mo2019}. We report $AP_{25}$, $AP_{75}$, and s-$AP_{50}$ on three instance levels. Bold numbers are better.
    }
    \label{tab:partnet-othermetric} \vspace{-3mm}
\end{table*}

{\small
\bibliographystyle{ieee_fullname}
\bibliography{reference}
}

\end{document}